\documentclass{article}

\usepackage{arxiv}

\usepackage[utf8]{inputenc} 
\usepackage[T1]{fontenc}    
\usepackage{hyperref}       
\usepackage{url}            
\usepackage{booktabs}       
\usepackage{amsfonts}       
\usepackage{nicefrac}       
\usepackage{microtype}      
\usepackage{lipsum}
\usepackage{graphicx}
\usepackage{amsmath,amsfonts}
\usepackage{algorithmic}
\usepackage{algorithm}
\usepackage{array}
\usepackage{textcomp}
\usepackage{stfloats}
\usepackage{verbatim}
\usepackage{cite}
\usepackage{subfigure}
\usepackage{multirow}
\usepackage{threeparttable}
\usepackage{booktabs}
\usepackage{caption}
\usepackage{authblk} 
\graphicspath{ {./images/} }

\title{Subpixel Edge Localization Based on Converted Intensity Summation under Stable Edge Region}

\author[1,2]{Yingyuan Yang}
\author[3,*]{Guoyuan Liang}
\author[1,2]{Xianwen Wang}
\author[4]{Kaiming Wang}
\author[3]{Can Wang}
\author[3]{Xinyu Wu}

\affil[1]{Shenzhen Institutes of Advanced Technology, Chinese Academy of Sciences, Shenzhen 518055, China.}
\affil[2]{University of Chinese Academy of Sciences, Beijing 10049, China.}
\affil[3]{Guangdong Provincial Key Laboratory of Robotics and Intelligent System, Shenzhen Institutes of Advanced Technology, Chinese Academy of Sciences, Shenzhen 518055, China.}
\affil[4]{Guilin University of Technology, Guilin 541000, China.}

\begin{document}
\maketitle
\begin{abstract}
To satisfy the rigorous requirements of precise edge detection in critical high-accuracy measurements, this article proposes a series of efficient approaches for localizing subpixel edge.
In contrast to the fitting based methods, which consider pixel intensity as a sample value derived from a specific model. We take an innovative perspective by assuming that the intensity at the pixel level can be interpreted as a local integral mapping in the intensity model for subpixel localization. Consequently, we propose a straightforward subpixel edge localization method called Converted Intensity Summation (CIS). To address the limited robustness associated with focusing solely on the localization of individual edge points, a Stable Edge Region (SER) based algorithm is presented to alleviate local interference near edges. Given the observation that the consistency of edge statistics exists in the local region, the algorithm seeks correlated stable regions in the vicinity of edges to facilitate the acquisition of robust parameters and achieve higher precision positioning. In addition, an edge complement method based on extension-adjustment is also introduced to rectify the irregular edges through the efficient migration of SERs. A large number of experiments are conducted on both synthetic and real image datasets which cover common edge patterns as well as various real scenarios such as industrial PCB images, remote sensing and medical images. It is verified that CIS can achieve higher accuracy than the state-of-the-art method, while requiring less execution time. Moreover, by integrating SER into CIS, the proposed algorithm demonstrates excellent performance in further improving the anti-interference capability and positioning accuracy.
\end{abstract}


\section{Introduction}
Edge detection is a fundamental task in image processing, serving as an essential precursor for subsequent advanced applications, like object detection and image segmentation.
Specifically, efficient edge detectors such as Canny\cite{Canny1986ACA} have been developed as fundamental tools in a multitude of visual tasks. 
However, these methods fail to satisfy the sectors requiring high precision\cite{Zhang2019RockringDA,Lu2021AutomaticWM,Jing2022RecentAO,Liu2023ANS}, such as industrial inspection, remote sensing, and medical imaging. 

\begin{figure}[t]
\centering
\subfigure[]
{
    \begin{minipage}[b]{.2\linewidth}
    \centering
    \includegraphics[width=0.8\textwidth]{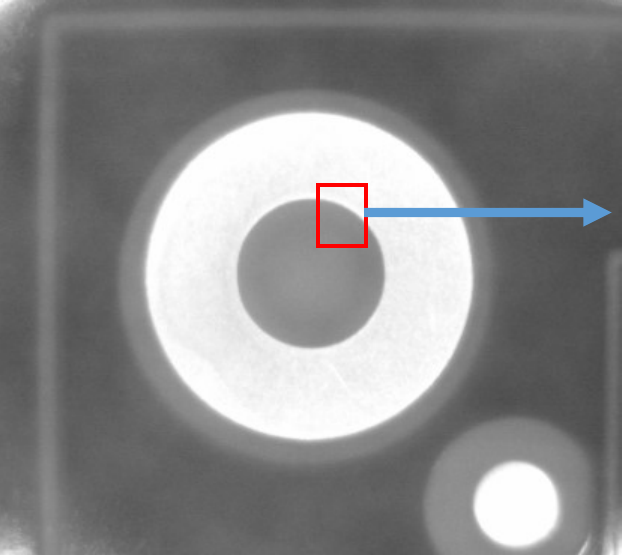} \\
    \vspace{0.2em}
    \includegraphics[width=0.8\textwidth]{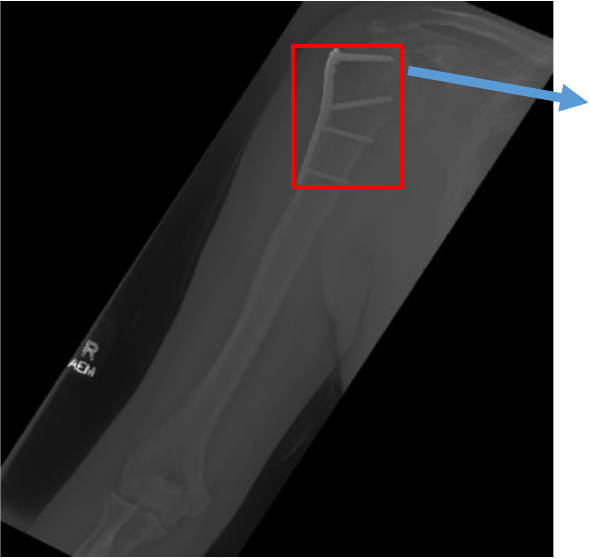} 
    \end{minipage}
}
\subfigure[]
{
  \begin{minipage}[b]{.2\linewidth}
    \centering
    \includegraphics[width=0.6\textwidth]{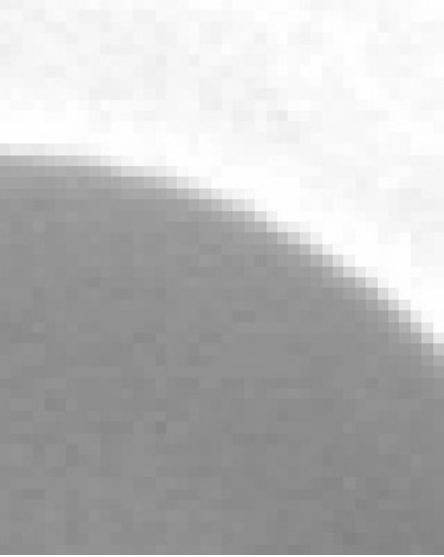} \\
    \vspace{0.2em}
    \includegraphics[width=0.6\textwidth]{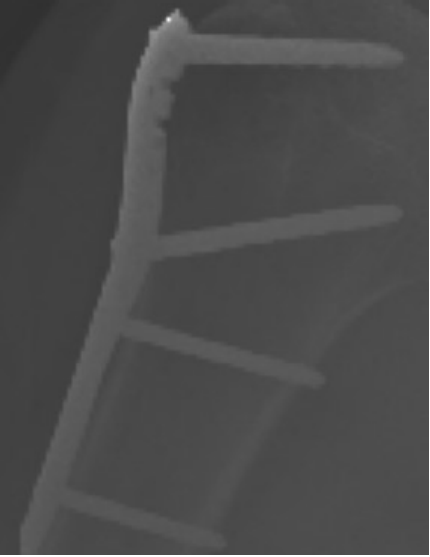} 
    \end{minipage}
}
\subfigure[]
{
  \begin{minipage}[b]{.2\linewidth}
    \centering
    \includegraphics[width=0.6\textwidth]{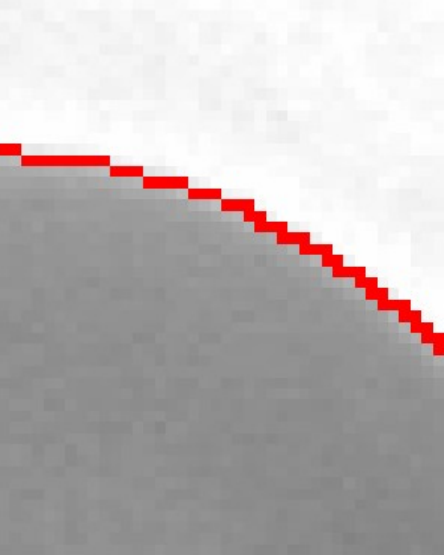} \\
    \vspace{0.2em}
    \includegraphics[width=0.6\textwidth]{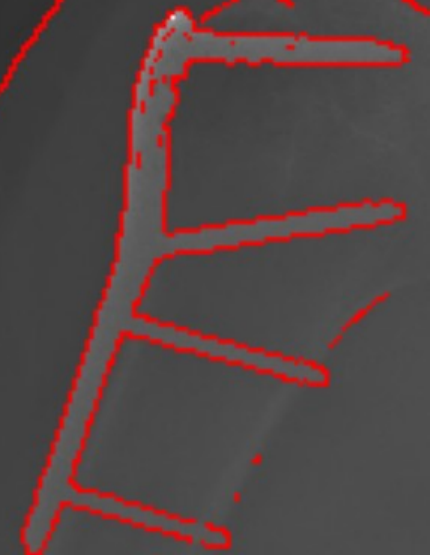} 
    \end{minipage}
    \label{fig1.1}
}
\subfigure[]
{
  \begin{minipage}[b]{.2\linewidth}
    \centering
    \includegraphics[width=0.6\textwidth]{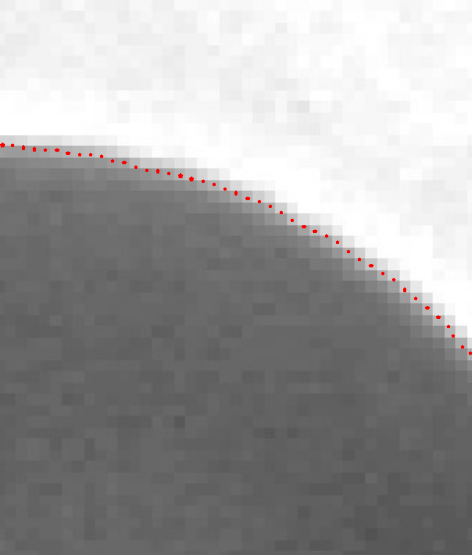} \\
    \vspace{0.2em}
    \includegraphics[width=0.6\textwidth]{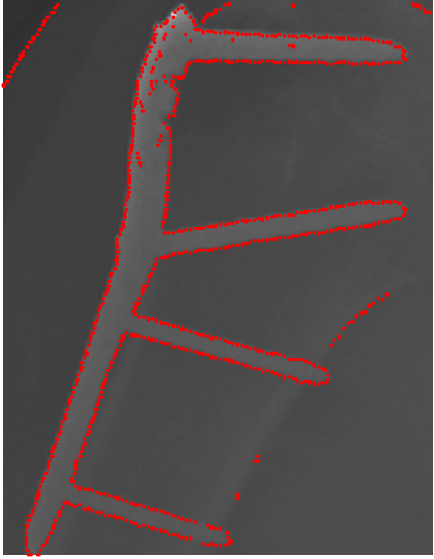} 
    \end{minipage}
    \label{fig1.2}
}
    \caption{The difference between edge detection and subpixel edge localization. (a) Original image. The top row is a  Printed Circuit Board (PCB) image and the bottom row is a medical image showing bone structure. (b) Enlarged sub-image bounded by the red rectangle. (c) Pixel level edge detection by Canny. (d) Subpixel edge localization by the proposed method.}
    \label{fig1}
\end{figure}

In response, researchers\cite{Hu2022CannySE} have extended the task of edge detection to the subpixel localization, with an emphasis on enhancing measurement accuracy, and the difference between them illustrated in Fig. \ref{fig1}.
Over the past decade, deep learning has shown considerable proficiency in semantic extraction. Consequently, the methods based on convolutional neural networks (CNN) \cite{Bertasius2014DeepEdgeAM,Su2021pixel,Zhou2023TheTB,zhou2024muge} have achieved remarkable success in identifying pixel-level edges with distinct boundaries. Nevertheless, these methods exhibit limitations in the precise location and measurement in subpixel level, and traditional methods offers a significant advantage in terms of sub-pixel localization.

In previous works, researchers have utilized the regularity of edges along gradient directions to design an elaborate edge model, typical techniques such as fitting\cite{YE2005453} and interpolation\cite{HERMOSILLA20081240}. In parallel, other works employ the relationship between a single edge pixel and its surrounding discrete region to accomplish precise positioning. Moment-based methods\cite{Kaur2011} utilize multiple order moment features to derive edge parameters crucial for localization. 
According to the observed dispersion of intensities and gradients, Seo\cite{Seo2018SubpixelEL} proposes a non-iterative method based on adaptive weighting of gradients to precisely estimate edge positions. 
And the partial area effect method \cite{TrujilloPino2013AccurateSE} incorporates the correlation between the area of a partial region and its intensity value, which serves as a constraint in fitting optimal edge curve.

\begin{figure}[tbp]
    \centering
    \includegraphics[width=3in]{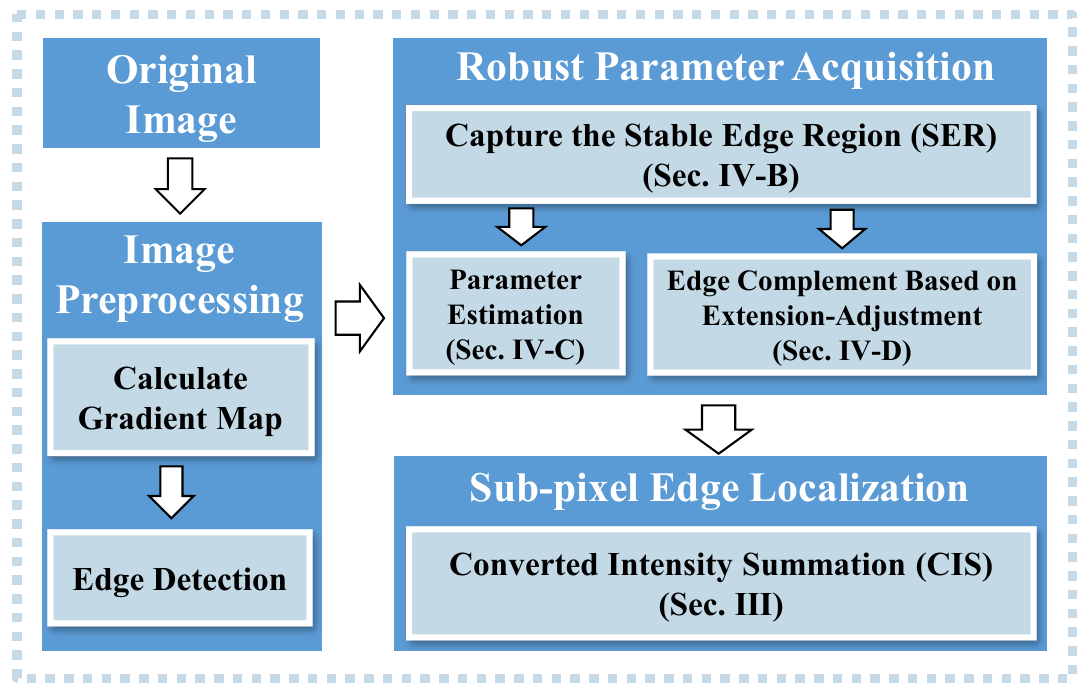} 
    \caption{The framework of the proposed method.}
    \label{fig2}
\end {figure}

In contrast to previous work that focuses on reconstructing edges using various models, our study explores associations between the intensity of a pixel and intensity model at subpixel level.
Traditionally, the intensity of a pixel has been treated as the discrete sample from a given model. However, in this paper, we bring up the idea that the intensity of a pixel can be viewed as an integral mapping over a specific region in a subpixel-level intensity model.
Drawing on the evolution of this concept, we implement a straightforward method called converted intensity summation (CIS) to achieve subpixel edge localization, by including the integral equivalence between the nonlinear symmetric curve and the step function.

In addition, previous research has primarily focused on the exact positioning of individual edge points while ignores correlations among them, which makes localization results heavy reliance on the results of edge detection and highly sensitive to local interferences. Upon observing that region adjacent to edges exhibits statistically consistent patterns, we present the concept of the stable edge region (SER). It is constructed by a stable local region near the edge pixels with the valid correlations, and used for reducing potential local interferences as well as capturing more robust statistical parameters. Furthermore, an edge complement method based on extension-adjustment is also proposed to correct irregular edge regions by efficient migration of parameters from SERs.

The overall framework of this work is shown in Fig. \ref{fig2}. The image first undergoes preprocessing to generate the gradient map and the initial edges are determined. Subsequently, the SER based algorithm is used to amalgamate stable regions proximate to edge points. And robust parameters are then derived from SER and the edge complement method. Finally, the proposed CIS method efficiently achieves accurate subpixel edge localization.

The main innovations of this work are delineated below:
\begin{enumerate}

\item A straightforward and effective subpixel edge localization method called converted intensity summation is proposed, which can achieve superior positioning accuracy compared to the state-of-the-art methods while vastly raising computational efficiency.

\item An algorithm based on stable edge region is proposed to enhance the robustness of subpixel edge localization, which captures stable statistical parameters by incorporating a broader range of related pixels in local region.

\item An edge complement method based on extension-adjustment is designed to correct the irregular edge region. By combination with SER, it bolsters the robustness of CIS, thereby further reducing the error in subpixel edge localization.

\end{enumerate}

The rest of the paper is organized as follows:
Sec. \ref{sec2} presents a review of related works, including subpixel edge localization and edge linking. 
Sec. \ref{sec3} describes a straightforward subpixel localization method based on converted intensity summation. 
Sec. \ref{sec4} provides the details of the stable edge region and the edge complement method with corresponding parameter estimation for CIS.
Sec. \ref{sec5} shows the experimental results and analysis on synthetic and real datasets.
A conclusion is given in Sec. \ref{sec6}.

\section{RELATED WORKS}\label{sec2}
\subsection{Subpixel edge localization}
Given the prevalence of subpixel edge detection in high-precision fields, a large number of related algorithms have been proposed. These algorithms may be broadly classified into three categories\cite{Fabijaska2015SubpixelED}: moment\cite{tabatabai1984edge,Ghosal1993OrthogonalMO,Bin2008SubpixelEL,Sun2014ARE}, interpolation\cite{Jensen,Qingli2003AIS,Chen201611,Single-Pixel-Multi-Point}, and fitting methods\cite{Nalwa1986699,Duan2018High,Liu2022ANS,2011Edge}.

Tabatabai and Mitchell \cite{tabatabai1984edge} developed the earliest moment-based method on subpixel edge localization, utilizing the relationship between the edge parameters and the gray moments (GMs). 
To obtain invariant under rotation, Ghosal et al.\cite{Ghosal1993OrthogonalMO} proposed Zernike orthogonal moments (ZOMs), but it cannot accurately detect small objects. The orthogonal Fourier–Mellin moments (OFMMs)\cite{Bin2008SubpixelEL} were proposed later to deal with this issue, and the testing accuracy was proven meeting the stringent requirements in medical image analysis or satellite remote sensing. 
Interpolation-based methods leverage interpolating the intensity or the corresponding derivative function of edge pixels to increase the edge information, thereby realizing subpixel edge detection. 
Jensen and Anastassiou\cite{Jensen} initially exploited non-linear interpolation to locate subpixel edges, their approach can produce noticeably sharper edges and exhibit a lower error than linear methods. 
Subsequent researchers designed different interpolation schemes\cite{Qingli2003AIS,Chen201611,Single-Pixel-Multi-Point}. 
The fitting-based methods usually assume that the curve of edge intensity or gradient variation follows to a functional model, after which subpixel coordinates are obtain from fitting curve using the least square method. 
Nalwa and Binford \cite{Nalwa1986699} proposed the first fitting-based method with the hyperbolic tangent function as the model. A blurred edge model \cite{YE2005453} was adopted to locate subpixel coordinates as well, and the experiments have demonstrated its good performance on both robustness and accuracy.
Similarly, Hagara and Kulla\cite{2011Edge} proposed the Erf function to approximate the true edge. Their method is considered to be the most accurate with the high computational cost. 

Recently, researchers\cite{Gioi2017ASE,Seo2018SubpixelEL,TrujilloPino2013AccurateSE,Chu2020SubpixelDM} have made remarkable progresses on methods other than that from above three traditional categories.
The method based on the partial area effect\cite{TrujilloPino2013AccurateSE} has been applied in many different fields\cite{LI2022Visual,Lu2022ANS,poyraz2024sub}, which assumes a valid intensity equation by using the discrete character and regional correlation of edge pixels.
Gioi\cite{Gioi2017ASE} incorporated the classic Canny and Devernay, but using only three pixels makes this approach sensitive to noises. Besides, Seo\cite{Seo2018SubpixelEL} presented a subpixel edge localization method based on the adaptive weighting of gradients (AWG). It has less computational cost and more accuracy than the Erf-fitting method.
And the algorithm based on the intensity integration threshold (IIT)\cite{Chu2020SubpixelDM} locates the subpixel point where the intensity integration reaches the threshold. 
In contrast to these studies, our research delves deeply into the relationship between intensity mapping at subpixel and pixel levels, which leads us to propose a more efficient and streamlined formula.

\subsection{Edge linking}
The following literatures \cite{XIE1992647, FaragandDelp,Ghita2002479,Wang,Topal2012EdgeDA,Akinlar2016PELAP,Seo2019SubpixelLL} have presented the representative works so far on edge linking at pixel level, which inspired our related research at subpixel level.

As one of the earliest researchers presenting the methods of linking edge pixels, Xie\cite{XIE1992647} proposed the concepts of horizontal edge elements and causal neighborhood windows to realize edge linking. 
Farag and Delp\cite{FaragandDelp} used the path metric based on the linear model as well as the A* algorithm to construct a new linking algorithm. Wang and Zhang \cite{Wang} improved previous linking method by calculating the edge direction within a specific local neighborhood and measuring the geodesic distance. After that, Akinlar and Chome\cite{Topal2012EdgeDA,Akinlar2016PELAP} introduced smart routing (SR) and predictive edge linking (PEL) to link neighboring anchors and obtain continuous edges.

\begin{figure}[t]
    \centering
    \includegraphics[width=2in]{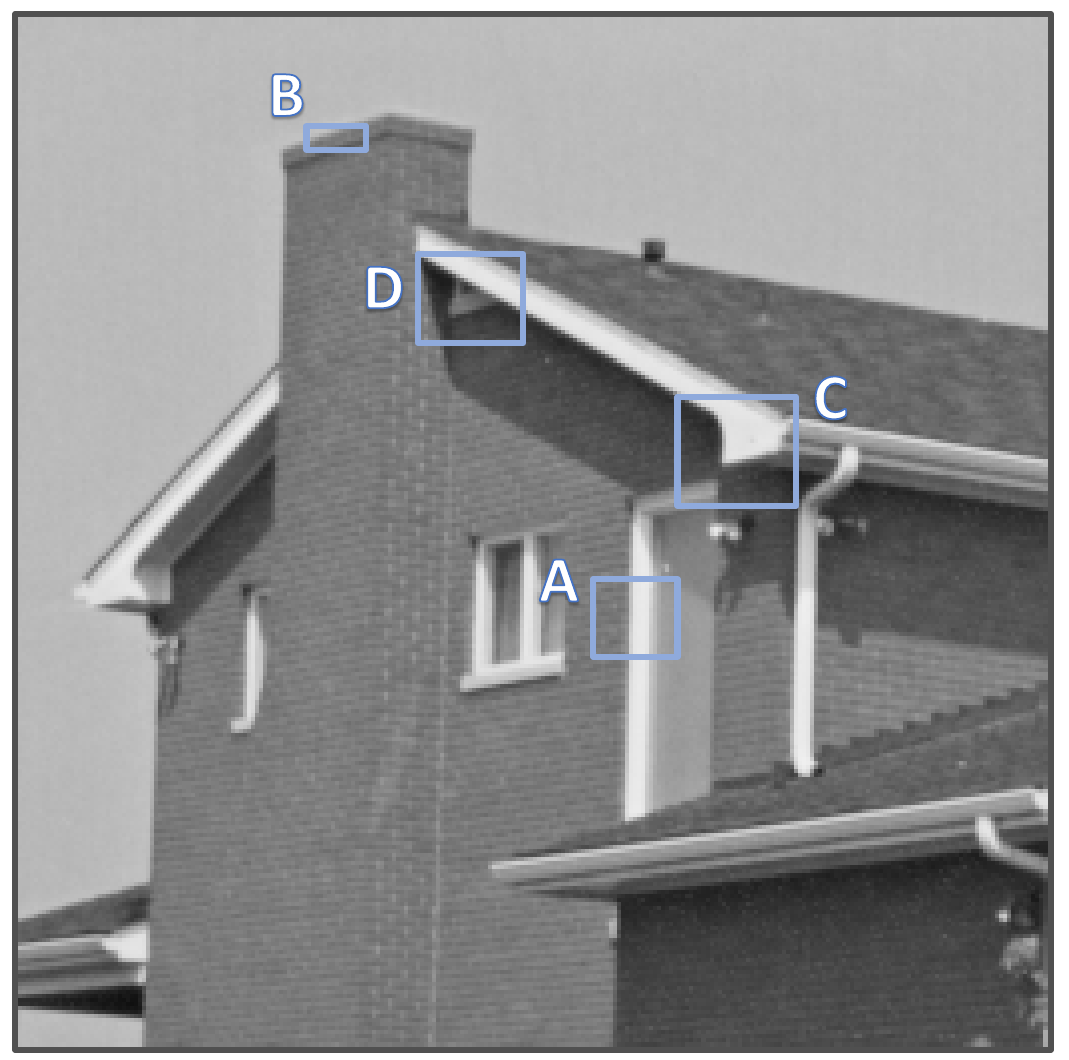} 
    \caption{Regions containing edges in the test image of a house.}
    \label{fig_s}
\end {figure}

\begin{figure}[h]
   \centering
    \subfigure{
    \includegraphics[width=0.7\linewidth]{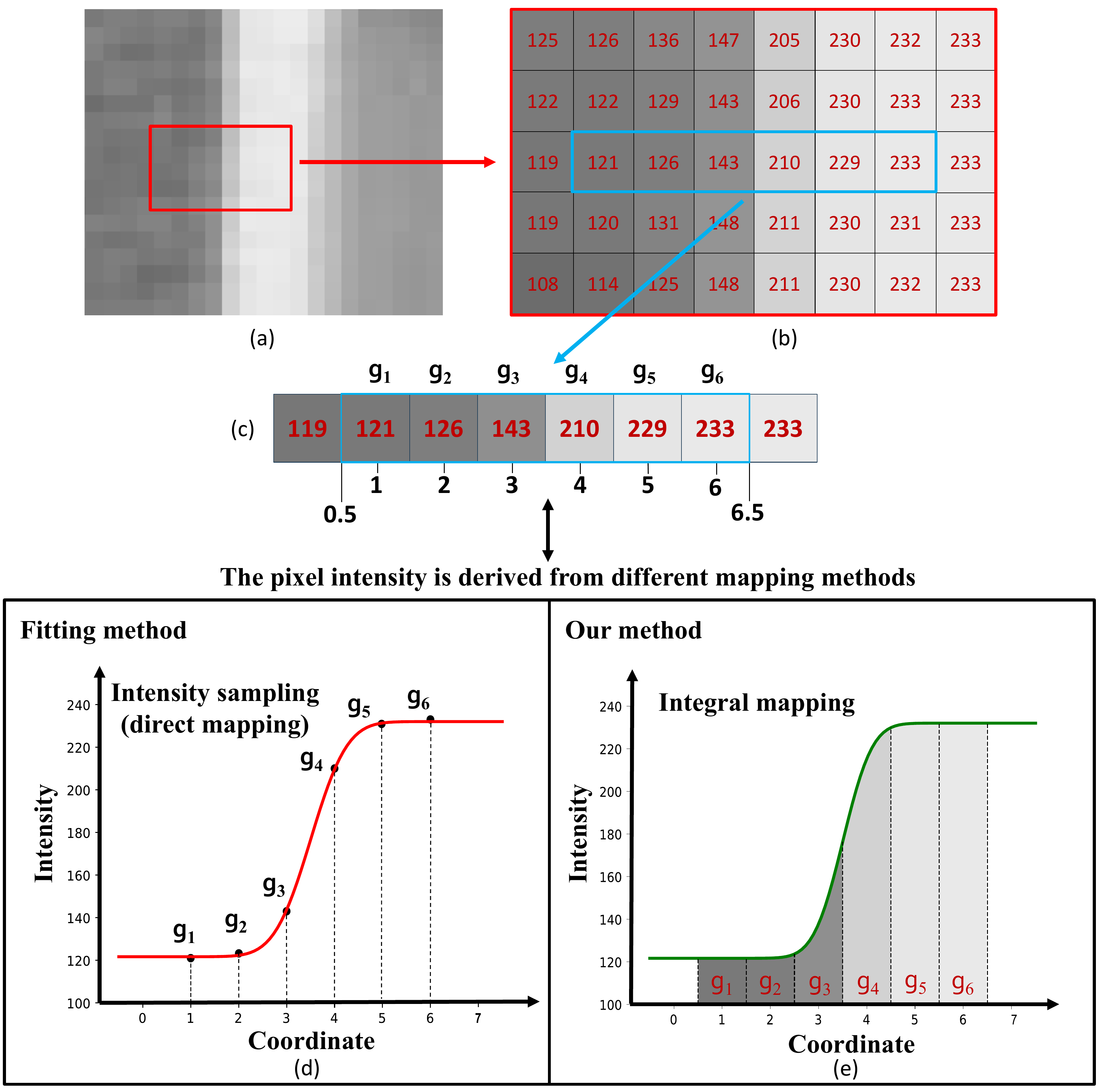}}
    \caption{The different intensity mappings in edge intensity curve. (a) Enlarged sub-image A form Fig. \ref{fig_s}. (b) Intensity distribution near an edge running vertically. (c) A sequence of $n$ pixels extending along the gradient orientation. 
    (d) Intensity sampling (direct mapping) used in the fitting method. (e) Integral mapping used in our method.}  
    \label{fig5}
\end{figure}

\begin{figure}[b]
\begin{minipage}[t!]{0.45\linewidth}
    \centering
    \includegraphics[width=2in]{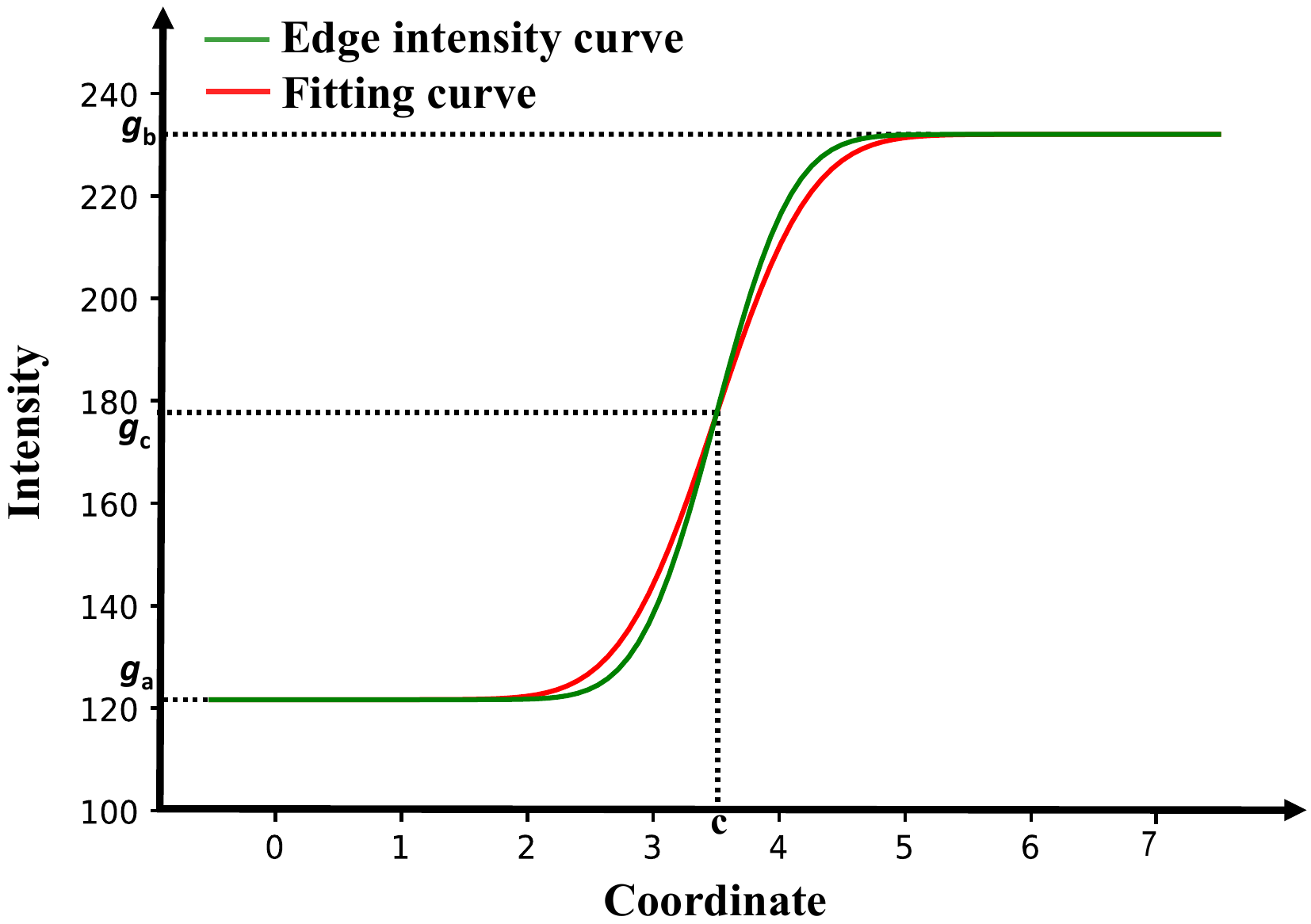} 
    \caption{The edge intensity curve and the curve from fitting pixel intensity. $c$ is the subpixel coordinate of the edge, while $g_c$ is its intensity. $g_a$ and $g_b$ are the intensities located on the either side of the edge.}
    \label{fig4}
\end{minipage}%
\hspace{5mm}
\begin{minipage}[t!]{0.45\linewidth}
    \centering
	\subfigure[]{
	\includegraphics[width=1.35in]{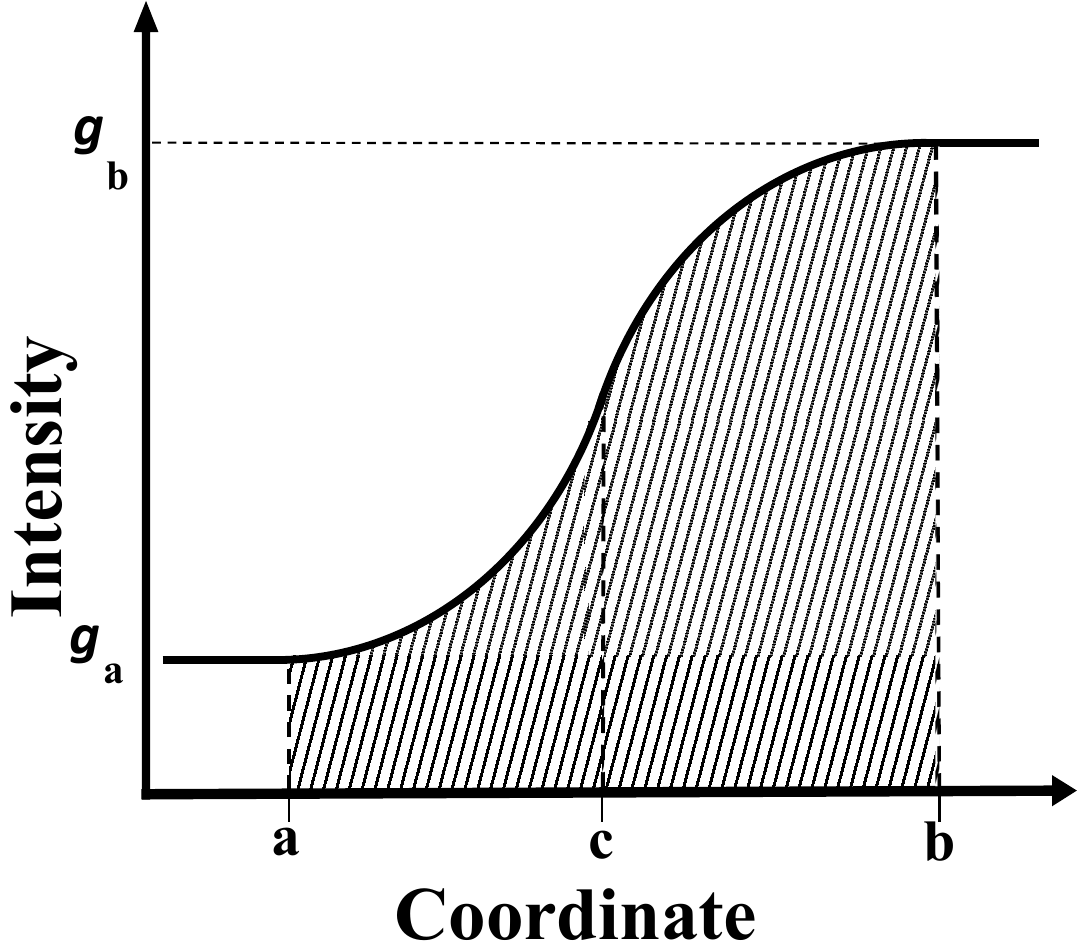} 
    \label{fig3.1}
    }
	\subfigure[]{
	\includegraphics[width=1.35in]{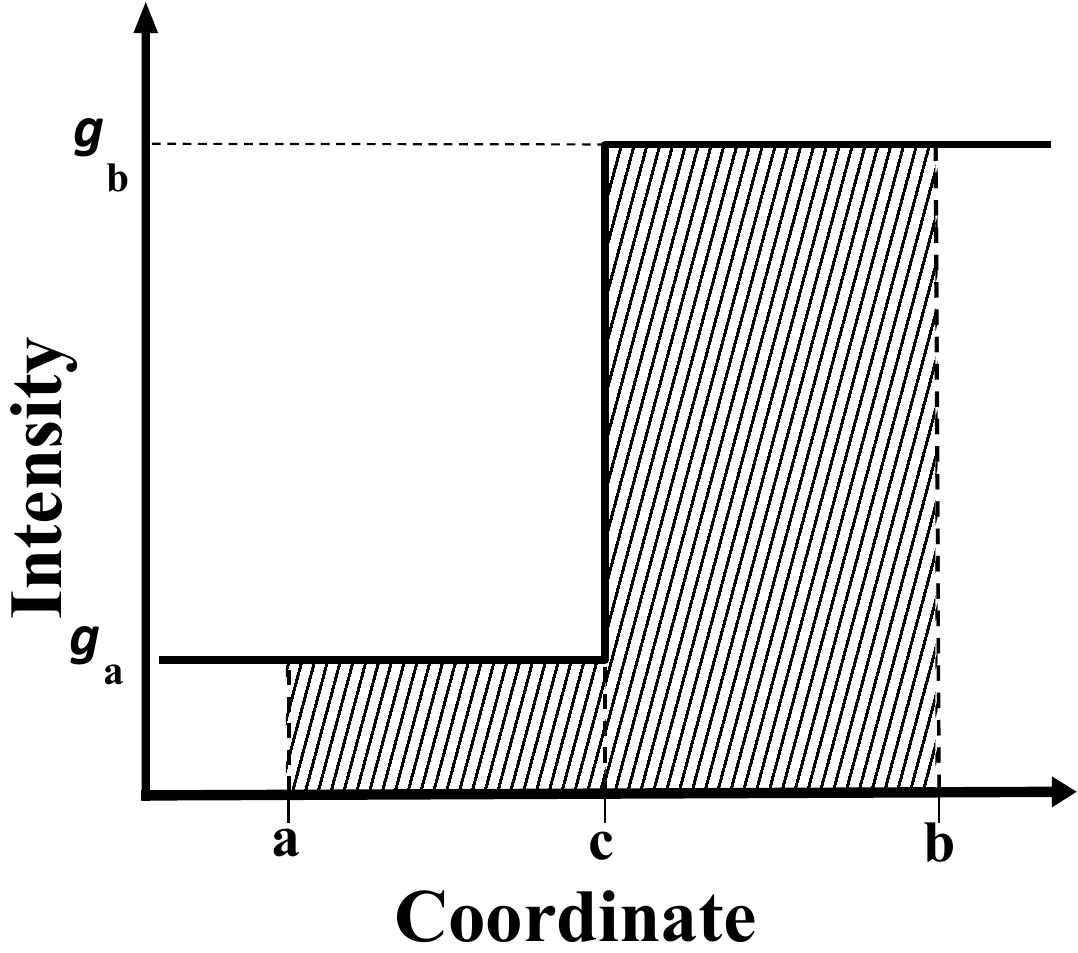} 
	\label{fig3.2}
    }	
	\caption{The integral in different functions. (a) The integral in the edge intensity curve. (b) The integral in the step function.}
    \label{fig3}
\end{minipage}

\end{figure}

\section{converted intensity summation}\label{sec3}
The fitting method can achieve the highest positioning accuracy, but it requires excessive computations.
In this section, we propose a straightforward yet competitive method for subpixel edge localization, called converted intensity summation (CIS).
Sec. \ref{sec3.1} delineates a novel theoretical hypothesis about integral mapping in the edge intensity curve. Sec. \ref{sec3.2} provides a simple formula for accurate subpixel edge localization. To more effectively illustrate our methods, we have selected a subset of local regions containing edges from Fig. \ref{fig_s} for testing.

\subsection{Integral mapping in the edge intensity curve}\label{sec3.1}

In fitting-based methods\cite{Nalwa1986699,Duan2018High,Liu2022ANS,2011Edge}, the subpixel coordinates are derived by fitting intensities at pixel locations to a designated curve  along the gradient direction of the edge.
Unlike the step function, this curve consists of a gradual symmetric region bounded by flat regions on both sides, as shown in Fig. \ref{fig5}(c). 
To aid in the analysis, we refer to this curve as the edge intensity curve.

Previous studies treat the intensity of a pixel as a sampling value on the edge intensity curve with a one-pixel length interval.
Contrary to the above viewpoint, Hermosilla et al.\cite{HERMOSILLA20081240} assert that the intensity of a pixel is derived from the mean value of all subpixel points within it. Building upon this premise, we propose the following assumption:

\textbf{Assumption 1}: The intensity at the subpixel level manifests as an edge intensity curve, whereas the intensity of a pixel can be defined by the integral mapping within a specified range of this curve.
Generally, for a pixel located at $p$, the integral interval is defined as $(p-0.5,p+0.5)$.

Fig. \ref{fig5} exhibits the difference between our hypothesis and the fitting method.
In the fitting method depicted in Fig. \ref{fig5}(c), pixel intensities are directly sampled from the edge intensity curve. Conversely, the method illustrated in Fig. \ref{fig5}(d) derives pixel intensities through the integration of local regions in the edge intensity curve.
It is evident that the interpretation of pixel intensities in our method differs from previous works, despite the similar curve shape.

Fig. \ref{fig4} illustrates the edge intensity curve and the fitting curve in the same coordinate system. 
It is noticed that the symmetry center and the intensities of smooth sides can remain constant in these curves. It validates that the hypothesis has consistent edge positioning points with the fitting method.

\subsection{Subpixel edge localization based on converted intensity summation}\label{sec3.2}
According to the analysis presented above, the summation of pixel intensities from 1 to $n$ in a sequence, is equivalent to the integral over the the edge intensity curve within the range of $(0.5,n+0.5)$, as formulized by:

\begin{equation}
    \begin{split}
    I=&\sum_{i=1}^ng_i=\int_{0.5}^{n+0.5}f(x)dx\\
    =&(a-0.5)*g_a+\int_{a}^{b}f(x)dx+(n+0.5-b)*g_b
    \end{split}
    \label{eq1}
\end{equation}
where $f(x)$ is the intensity curve, as shown in Fig. \ref{fig3.1}. $a$ and $b$ are the are the coordinates of the two ends of the nonlinear region, and $g_a$ and $g_b$ are the intensities of the smooth regions, respectively.

To simplify the integral calculation in Eq. (\ref{eq1}), a step function is constructed whose abrupt point locates at the edge center $c$, with the lower and upper boundary aligned to the intensities $g_a$ and $g_b$, as illustrated in Fig. \ref{fig3.2}.

Owing to the inherent symmetry, the integral along the curve from $a$ to $b$ is equivalent to that of the step function, which can be formulized as:
\begin{equation}
\label{eq2}
\int_{a}^{b}f(x) dx= (c-a)*g_a+(b-c)*g_b 
\end{equation}
where $c$ is the abrupt point, representing the subpixel edge position. And the Eq. (\ref{eq7}) can be reformulated as follows:
\begin{equation}
\label{eq3}
\begin{split}
I=(c-0.5)*g_a+(n+0.5-c)*g_b
\end{split}
\end{equation}
The subpixel coordinate of the edge is:
\begin{equation}
\label{eq4}
c=\frac{I-n*g_b}{g_a-g_b}-\frac12
\end{equation}

For a given sequence, parameters $I$ and $n$ can be directly determined. In simple calculations, $g_a$ and $g_b$ can be computed as the average intensity of the region that exhibits smoothness without significant intensity variation. The experiments in Section \ref{sec5.1}, prove that CIS offers superior accuracy compared to the fitting method, while also significantly reducing computational time.

\section{Stable edge region}\label{sec4}

Despite the superior performance and efficiency offered by CIS, it exhibits a notable limitation similar to the previous methods. Specifically, it focuses solely on individual edge points and heavily relies on the results of edge detection, making it highly susceptible to interference.
To this end, the section introduces the stable edge region which incorporates a broader range of related pixels to capture stable statistical parameters, thereby enhancing the robustness of subpixel edge localization.

Sec. \ref{sec4.1} and Sec. \ref{sec4.2} delineate the relevant theory and the determination of SER, respectively. 
Sec. \ref{sec4.3} provides the parameter estimation for CIS on SER. 
For complex regions hard to obtain stable parameters, we propose an Extension-Adjustment algorithm in Sec. \ref{sec4.4} to rectify the irregular edges.

\begin{figure}[t]
	\centering
	\subfigure[]{
		\includegraphics[width=0.35\linewidth]{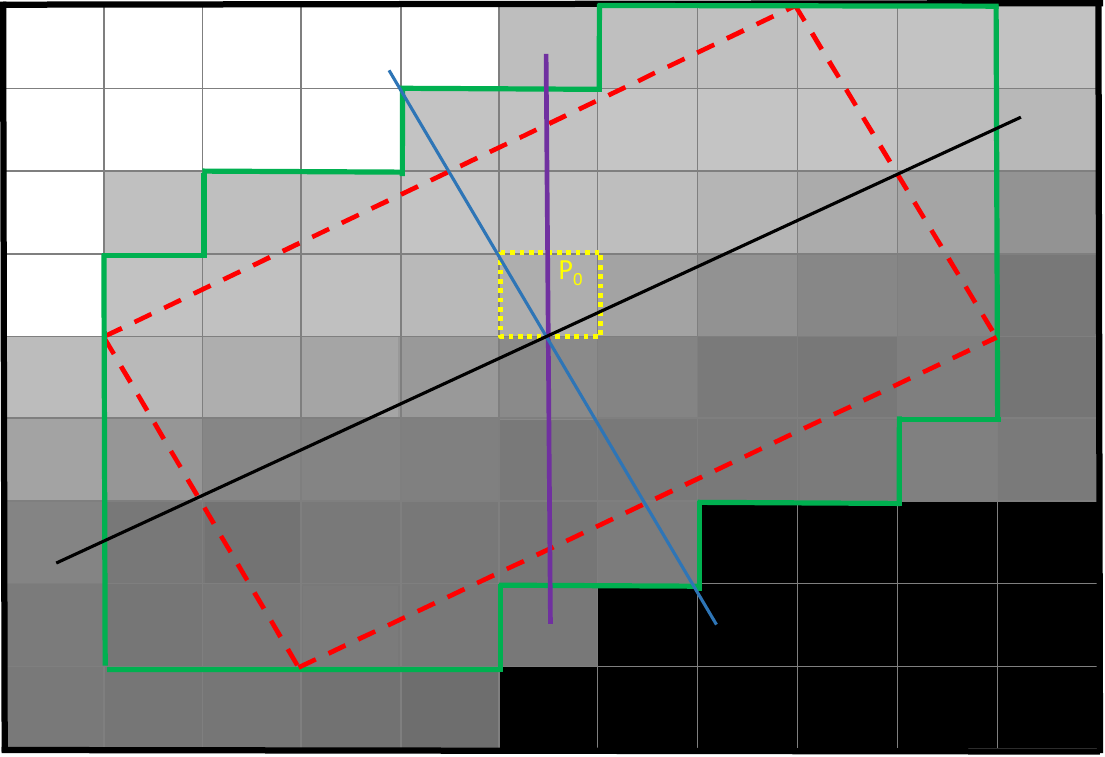} }
	\subfigure[]{
		\includegraphics[width=0.35\linewidth]{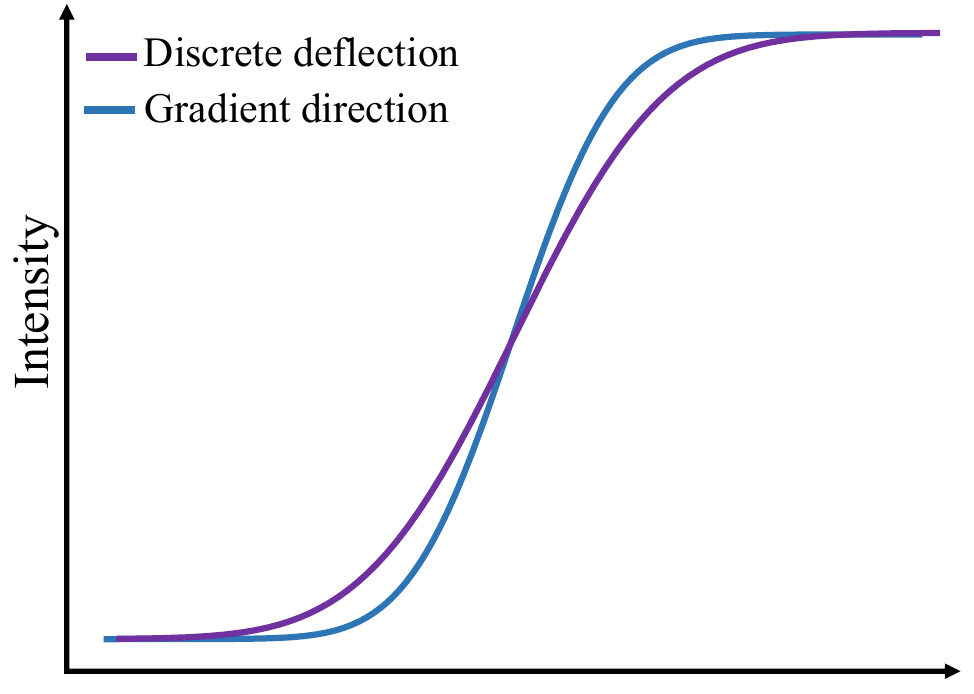}}
	\caption{A local edge region and the intensity curves. (a) Enlarged sub-image B form Fig. \ref{fig_s}. The region bounded by a red rectangle, represents a local area along the directions of the edge gradient and tangent. The black line denotes the edge direction, while the blue line and purple line represent the gradient direction and DD respectively. The pixels encompassed by green lines are those associated with the red rectangle. (b) The edge intensity curve along the gradient direction and DD.}
	\label{fig6}
\end{figure}

\begin{figure}[b]
	\centering
	\subfigure[]
	{\rotatebox{90}{\scriptsize{\textbf{~~~~~~~~~~Complex~~~~~~~~~~~~~~~Steep~~~~~~~~~~~~~~~~~Simple}}}
		\rotatebox{90}{\scriptsize{\textbf{~~~~~~~~~~~~~~edge~~~~~~~~~~~~~~~~~~edge~~~~~~~~~~~~~~~~~~~edge}}} 
		\begin{minipage}[b]{.3\linewidth}
			\centering
			\includegraphics[width=1\textwidth]{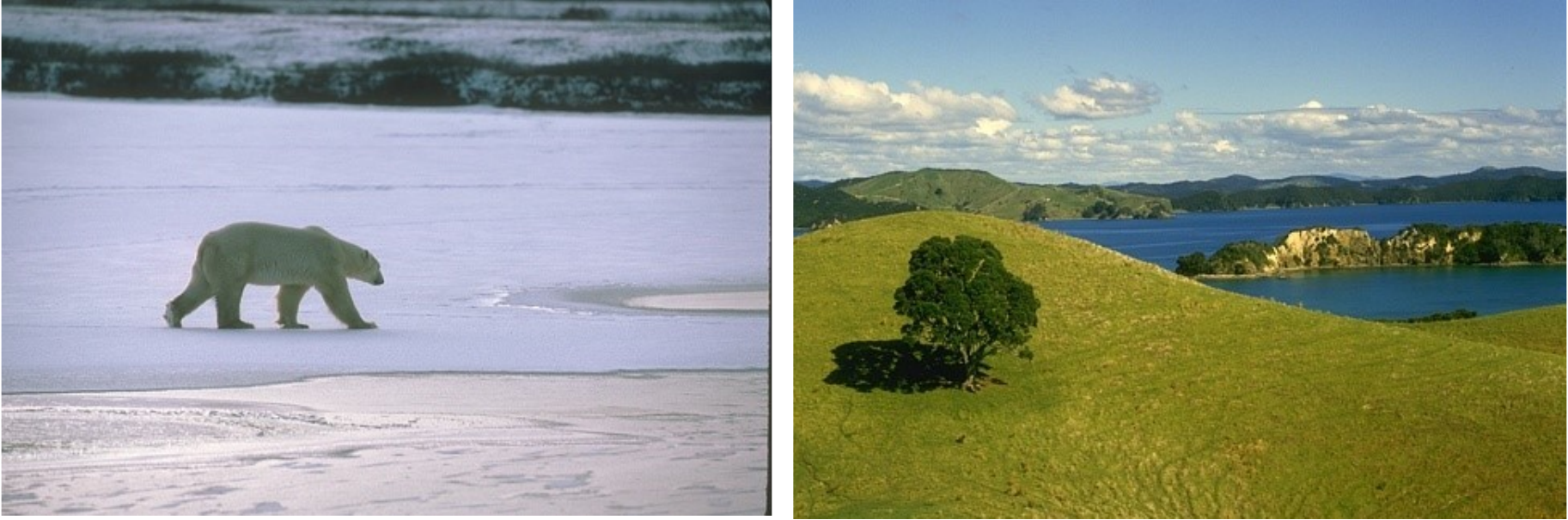} \\
			\vspace{-1.9em}
			\begin{center}
				\scriptsize  \quad
			\end{center} 
			\includegraphics[width=1\textwidth]{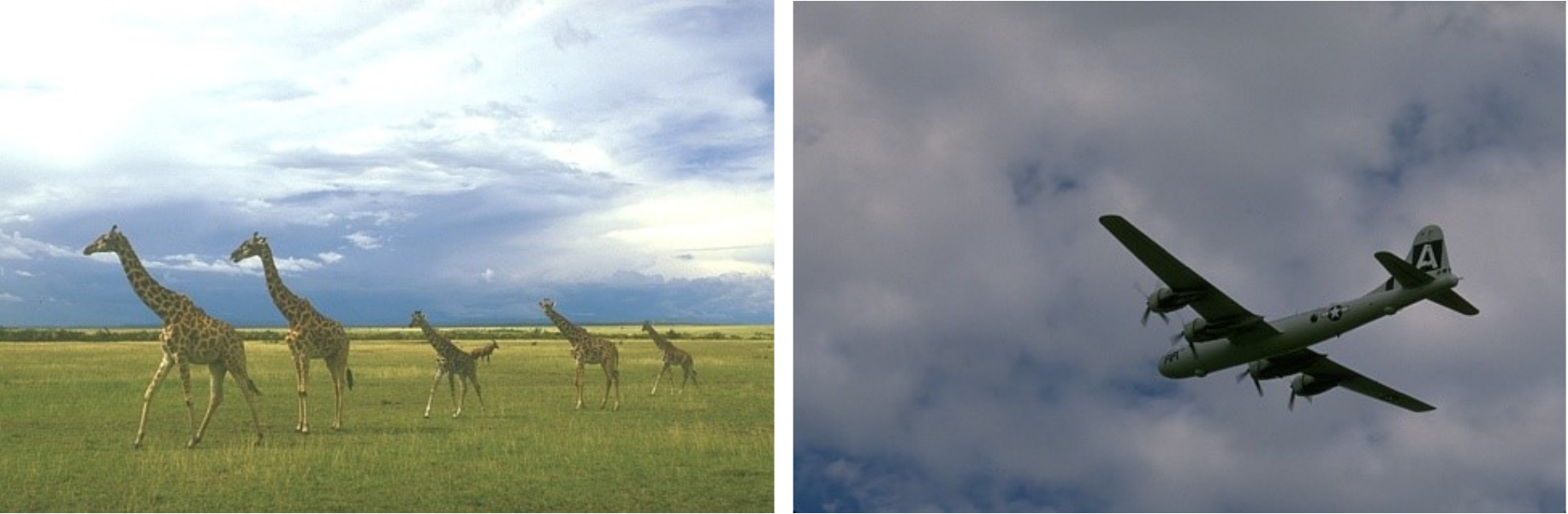} \\
			\vspace{-1.9em}
			\begin{center}
				\scriptsize  \quad
			\end{center} 
			\includegraphics[width=1\textwidth]{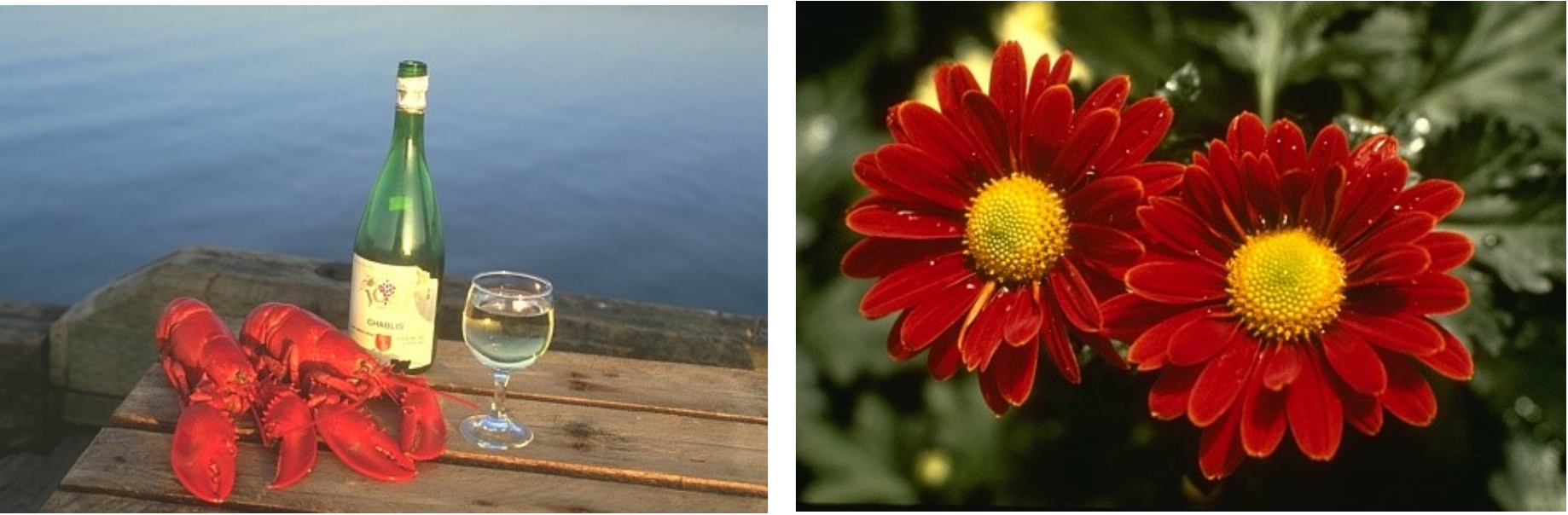} 
			\vspace{-1.5em}
			\begin{center}
				\scriptsize \quad
			\end{center} 
		\end{minipage}
	}
	\subfigure[]
	{
		\begin{minipage}[b]{.3\linewidth}
			\centering
			\includegraphics[width=1\textwidth]{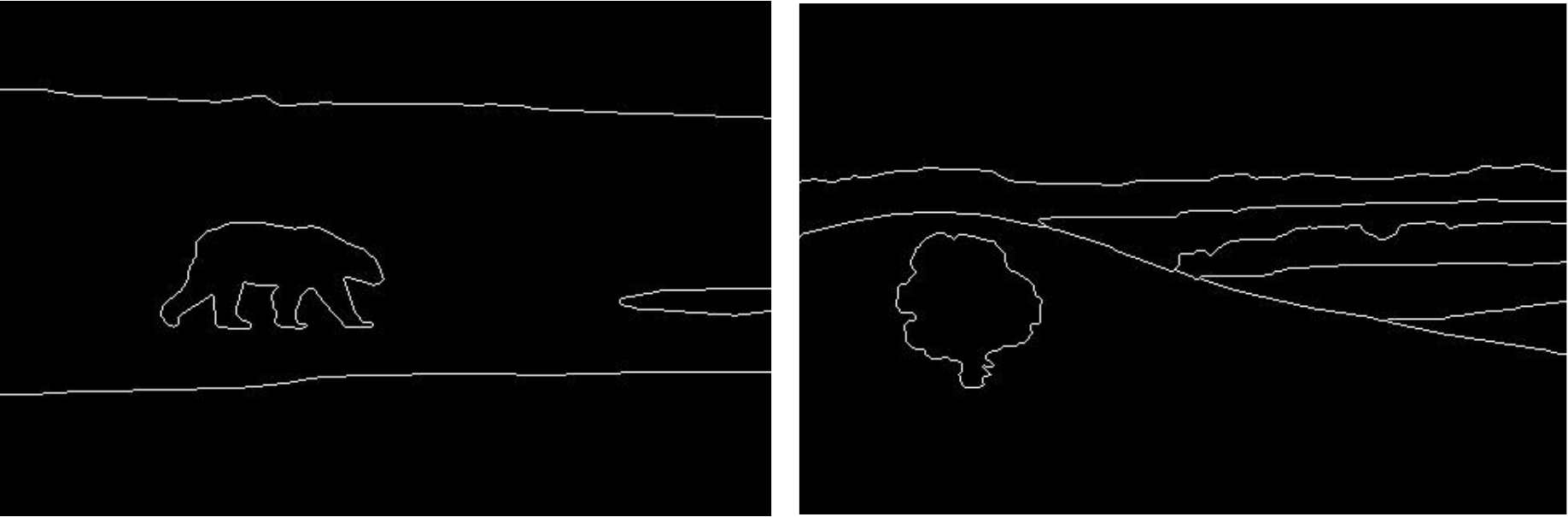} \\
			\vspace{-1.9em}
			\begin{center}
				\scriptsize  \quad
			\end{center} 
			\includegraphics[width=1\textwidth]{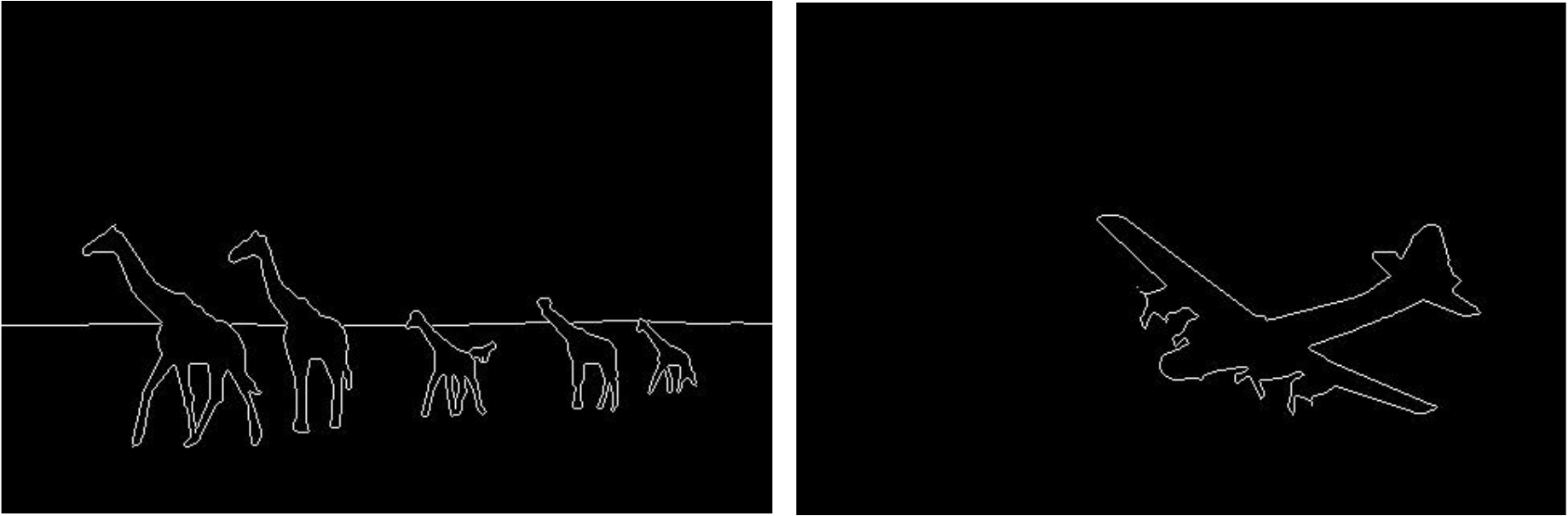} \\
			\vspace{-1.9em}
			\begin{center}
				\scriptsize  \quad
			\end{center} 
			\includegraphics[width=1\textwidth]{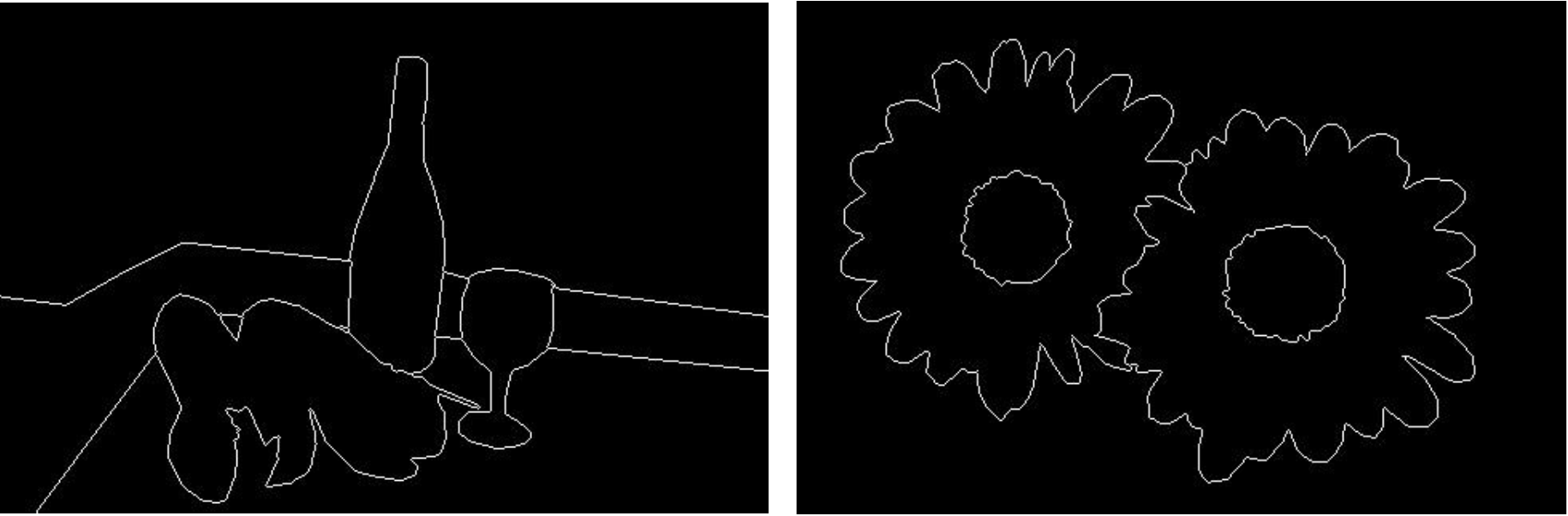} 
			\vspace{-1.5em}
			\begin{center}
				\scriptsize  \quad
			\end{center} 
		\end{minipage}
	}
	\caption{A subset from BSDS500. (a) Original image. (b) Extracted edges.
	The first row depicts pictures with simple edges, which consist of curves and partial corners; the second row includes some steep edges; the final row displays the edges with a large number of crossings under complex backgrounds.}
	\label{fig7}
	
\end{figure}

\subsection{Relevant theory for stable edge region}\label{sec4.1}

In accordance with the fitting method, the intensity distribution along the horizontal or vertical direction also can match the edge intensity curve.
The horizontal or vertical direction in close proximity to the gradient direction is designated as the Discrete Deflection (DD) in this paper. Fig. \ref{fig6} displays the intensity curves along the gradient direction and DD.

The sequence of pixels containing the edge intensity curve along the DD is defined as the Discrete Deflective Sequence (DDS). 
Given the similarity exhibited in adjacent edge pixels in the tangential direction, the following assumption is deduced:

\textbf{Assumption 2}: Within the local region traversed by an edge, each DDS can maintain the consistency of the edge intensity curve. In detail, they have similar intensity in the smooth region and similar length in the nonlinear region.

\textbf{Stable edge region (SER)} is defined as a local region that adheres to consistency in Assumption 2.

To validate the assumption and universality of SER existence, we randomly pick up several images with different levels of edge complexity from the BSDS500 dataset\cite{Arbelez2011ContourDA}, as shown in Fig. \ref{fig7}.
The specific calculation steps are as follows:
Firstly, we extract the local regions traversed by the edge using a 7*7 window,and obtain several DDSs consisting of seven pixels in these regions. After that, we calculate the proportion of cosine similarity greater than 0.9 between each pair of DDSs in the local region. Regions where this proportion surpasses 95\% are deemed to nearly meet Assumption 2, and the ratio of these eligible regions in the entire image is subsequently calculated.
As presented in Table \ref{tab1}, the majority of local regions exhibit the consistency among DDSs, except regions with high curvature or strong interference.

\begin{table*}[t!]
	\centering
	\begin{threeparttable}
		\newcommand{\tabincell}[2]{\begin{tabular}{@{}#1@{}}#2\end{tabular}}
		\caption{The statistical results for verification}
		\label{tab1}
		\renewcommand\arraystretch{1.3}
		\begin{tabular}{ccccc}
			\hline
			\tabincell{c}{The type \\ of edge} 
			& \tabincell{c}{The total number \\ of edge pixel} 
			& \tabincell{c}{The number of the local \\region containing edges}
			& \tabincell{c}{The number of the \\ local region with \textit{P}$>$95\%}
			& \tabincell{c}{The ratio of the \\ local region with \textit{P}$>$95\%}\\ 
			\hline
			Simple edge  & 3828  & 880 & 854 & 97.04\% \\    
			Steep edge   & 3131  & 707 & 672 & 95.05\% \\ 
			Complex edge & 4402  & 1025 & 933 & 91.03\% \\
			\hline
		\end{tabular}
		\begin{tablenotes}    
			\footnotesize 
			\item $P$: the proportion of cosine similarity greater than 0.9 between each pair of DDSs in the local region.
		\end{tablenotes}     
	\end{threeparttable}
\end{table*}

\subsection{Algorithm for SER determination}\label{sec4.2}
The process of determining SER includes the following three steps:

\textbf{Step 1}. Pre-processing. 

\textbf{Step 2}. Acquire the stable DDS for qualified edge pixels.

\textbf{Step 3}. Expand the DDS along the tangent direction.

\subsubsection{Pre-process.}
The Sobel operator is employed to compute the vertical and horizontal gradient for each pixel, the edge pixels are subsequently extracted based on gradient thresholds and non-maximum suppression. The direction with higher gradient determined as the DD for each edge pixel. 

\begin{figure}[b]
\begin{minipage}[t]{0.45\linewidth}
	\centering
	\includegraphics[scale=0.4]{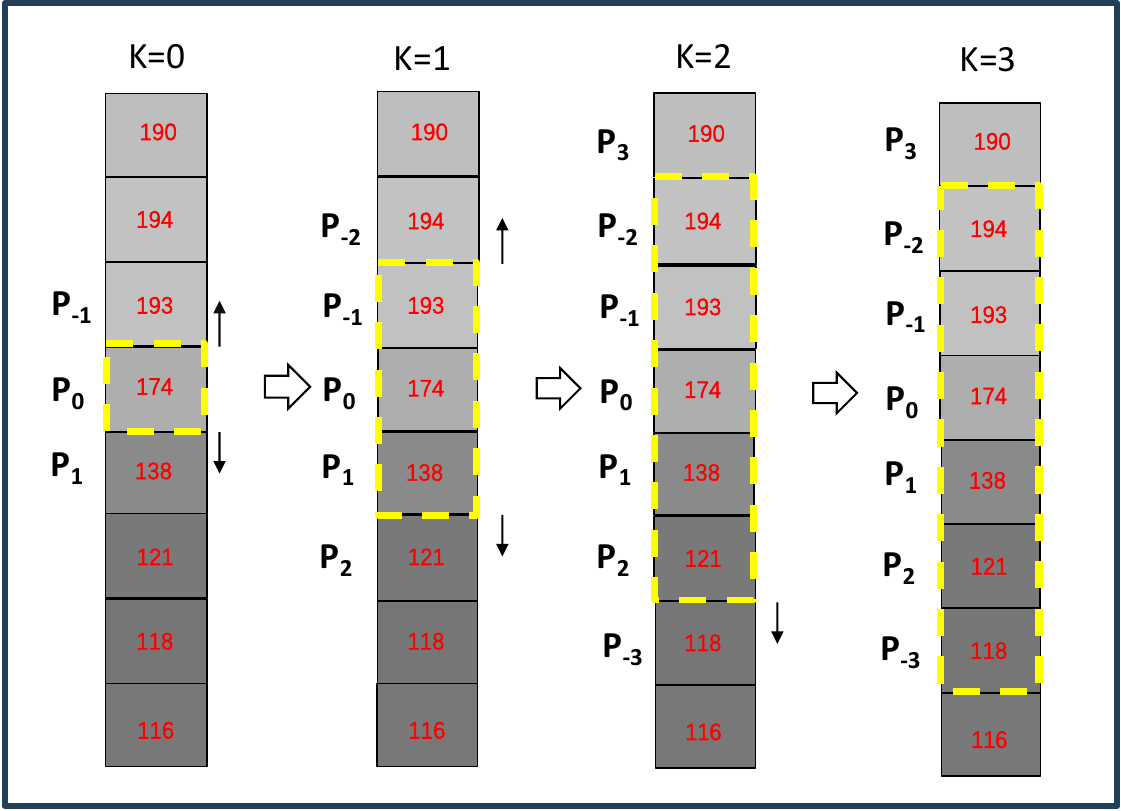}
	\caption{The process of acquiring the original stable DDS. The yellow rectangle stands for the current DDS, which expands in both directions of DD. In the second turn of expansion ($K = 2$), the top has reached the smooth region and the expansion stops, while the bottom continues to expand. Finally, when $K = 3$, both sides have successfully entered the smooth region, therefore the expansion process is completed.}
	\label{fig8}
 \end{minipage}%
 \hspace{5mm}
  \begin{minipage}[t]{0.45\linewidth}
	\centering
	\includegraphics[scale=0.39]{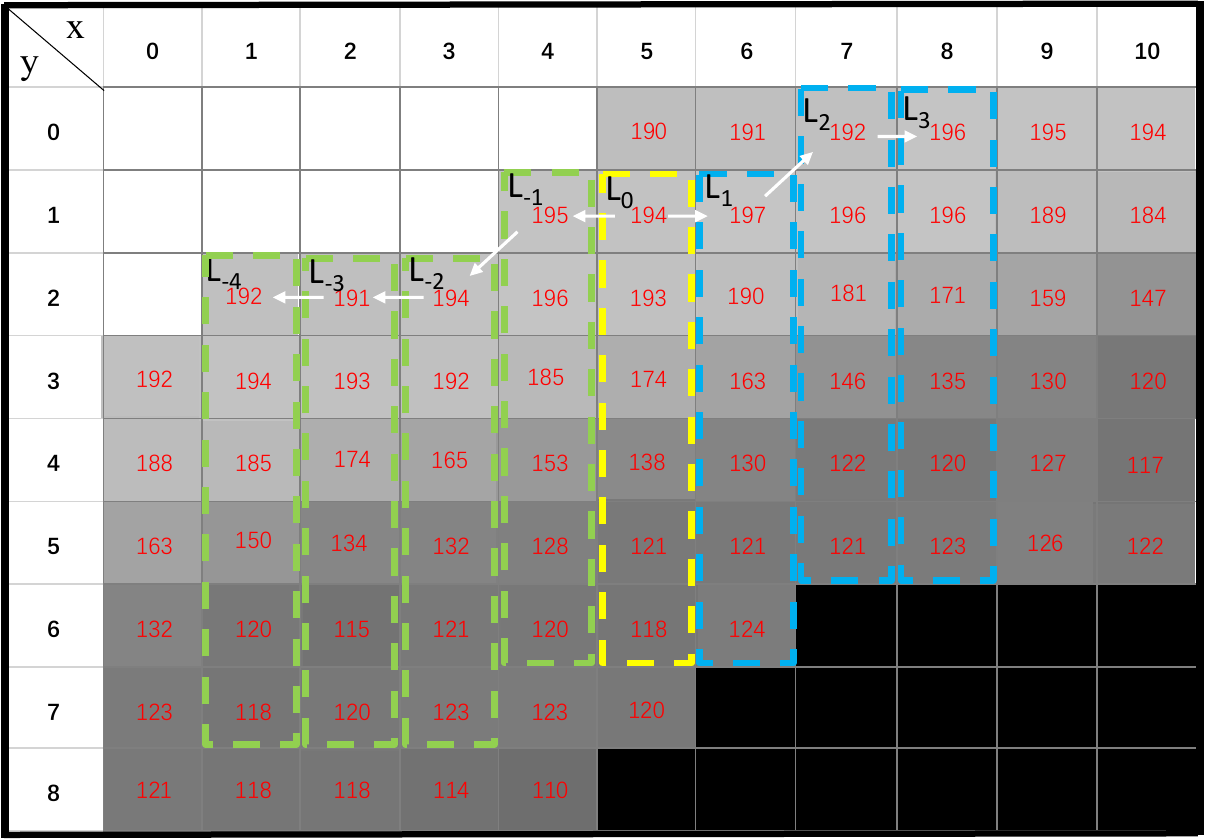}
	\caption{The process of expanding along the tangent direction in Fig. \ref{fig8}, with the left expansion represented in green and the right expansion in blue.} 
	\label{fig9}
	\end{minipage}
\end{figure}

\subsubsection{Acquire the stable DDS for qualified edge pixels}
In this step, we will browse all edge pixels and try to build the stable DDS for every qualified edge pixel by incrementally expanding pixels along both directions of DD. 
During this process, the mean intensity and directional derivative ratio are used for judging stability, and endmost variation of intensity is used for judging termination.

Given the symmetry of the intensity curve in DDS, the mean intensity in the whole sequence should remain consistent during the expansion process. And a directional derivative ratio can be served to describe the directional stability within the sequence. The formulas are as follows:
\begin{align}
	\label{eq5}
	m_k =\frac{1}{2k+1}\sum_{i=-k}^{k}g_i,\quad
	\theta_k =\frac{\sum_{i=-k_d}^{k_u}G_{yi}}{\sum_{i=-k_d}^{k_u}G_{xi}}
\end{align}
where $g_i$ is the intensity of pixel $i$, $k = max(k_u,k_d)$, signifies the maximum times of expansion, and $k_u$ and $k_d$ represent the number of pixels expanded in two directions of DD. $G_{yi}$ and $G_{xi}$ are the horizontal and vertical gradients at pixel $i$, respectively.
The stability constraint in $k$ step of expansion is calculated as follows:
\begin{equation}
\begin{aligned}
	\label{eq6}
	& \Delta_{s}^k = \min(\frac{|m_{k}-m_{k-1}|}{th_m},\frac{|\theta_{k}-\theta_{k-1}|}{th_\theta})\\
\end{aligned}
\end{equation}
where $th_m$ and $th_\theta$ are different predefined thresholds.
Additionally, the endmost variation of intensity is employed as a criterion to terminate the expansion, as formularized by: 
\begin{equation}
\begin{aligned}
	\label{eq7}
	& \Delta_{ev} = \max(|g_{-k_d}-g_{-k_d+1}|,| g_{k_u}-g_{k_u-1}|)
\end{aligned}
\end{equation}

All relevant parameters are recalculated after each round of expansion. 
If stability constraint $\Delta_{s}$ is larger than 1.0 during the expansion process, it is inferred that the disturbance has compromised the stability of the DDS. As a result, this edge pixel no longer meets the criteria for constructing a stable DDS and is therefore discarded.
If the above does not occur, the expansion will continue till $\Delta_{ev}$ falls below a predefined threshold $th_{ev}$. The termination of expansion indicates the DDS has covered the whole nonlinear region, and the current sequence is then added to a candidate set for subsequent analysis. Fig. \ref{fig8} illustrates the process of acquiring a DDS for edge pixel $p_0$ in Fig. \ref{fig6}(a).

\begin{figure}[b]
\centering
\begin{minipage}[b]{0.2\linewidth} 
    \centering
    \subfigure[]{
    \includegraphics[width=\textwidth]{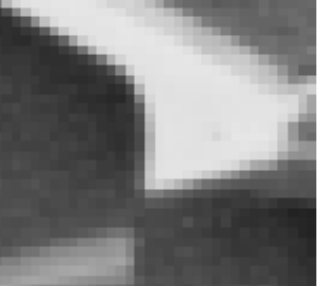}} 
    \end{minipage}
\hfill 
\begin{minipage}[b]{0.2\linewidth}
    \centering
    \subfigure[]{
    \includegraphics[width=\textwidth]{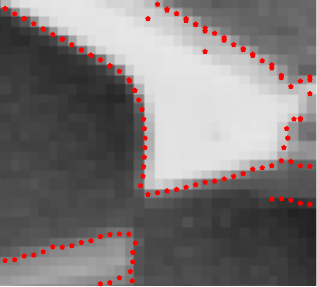}}
    \end{minipage}
\hfill
\begin{minipage}[b]{0.2\linewidth}
    \centering
    \subfigure[]{
    \includegraphics[width=\textwidth]{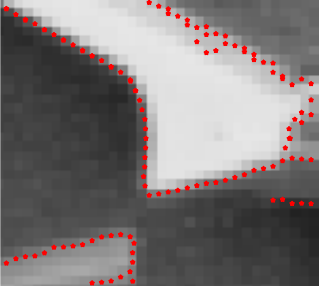}}
    \end{minipage}
\hfill
\begin{minipage}[b]{0.2\linewidth}
    \centering
    \subfigure[]{
    \includegraphics[width=\textwidth]{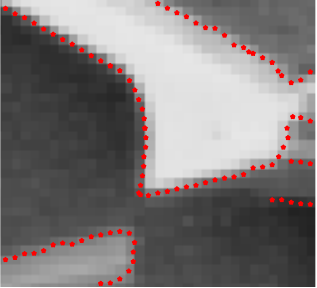}}
    \end{minipage}
\caption{The comparison of the localization for various methods. (a) The sub-image C from Fig. \ref{fig_s}. (b) Sub-pixel edge localization by fitting method based Erf \cite{2011Edge}. (c) Sub-pixel edge localization by CIS. (d) Sub-pixel edge localization by CIS + SER (without the edge complement), where outliers around edge region are properly cleaned.}
\label{figm2}
\end{figure}

\renewcommand{\algorithmicrequire}{\textbf{Input:}}
\renewcommand{\algorithmicensure}{\textbf{Output:}}
\begin{algorithm}[b!]
\caption{Parameter estimation for CIS in SER}
\begin{algorithmic}[1]
\REQUIRE $\mathbb{E}_S$: a pixel group in SER, $g_p$: the intensity of pixel $p$.
\ENSURE $g_a, g_b$: the intensities of smooth areas.
\STATE Calculate the mean $m_0$ and variance $v_0$ of $\mathbb{E}_S$.
\STATE Generate two pixel group on both sides of edges.\\ $\mathbb{E}_a\leftarrow \{p | g_i>m_0+v_0, p\in \mathbb{E}_S\},$ \\ $\mathbb{E}_b\leftarrow \{p| g_i<m_0-v_0, p\in \mathbb{E}_S\}.$
\STATE Obtain stable difference $D_0$ between the two groups.\\
$\mathbb{E}_D\leftarrow \{d | d=g_i-g_j, i\in \mathbb{E}_a, j\in \mathbb{E}_b\}$,\\
$D_0$ is the value with the highest frequency in $\mathbb{E}_D$.
\STATE Calculate the mean $m_a$ and variance $v_a$ of $\mathbb{E}_a$.
\STATE Calculate the mean $m_b$ and variance $v_b$ of $\mathbb{E}_b$.
\IF {$v_a < v_b$} \STATE $g_a = g_{a0},g_b=g_{a0}-D_0$
\ELSE \STATE $g_a = g_{b0}+D_0,g_b=g_{b0}$
\ENDIF
\end{algorithmic}
\end{algorithm}

\subsubsection{Expand the DDS along the tangent direction}
The subsequent step expands the acquired stable DDS along the tangent in both directions in turn. 
The expansion starts along the tangent towards both sides by one pixel each time.
Due to the consistency of SER, the length of each DDS along the tangent expansion remains unaltered. To make sure each DDS achieves optimal coverage over the nonlinear region of the edge intensity curve, the pixel situated on the expansion direction with the smallest intensity difference to the preceding edge pixel is chosen as the initial new DDS center. 
Finally, the expansion process stops based on a judgement derived from the relative derivative ratio and relative mean.

The directional derivative ratio (Eq. (\ref{eq5})) is also imposed to prevent outliers caused by high curvature, denoted by $\theta_l$, $l$ is $l$-th expanded DDS.
And given the uniform distribution of both foreground and background intensity in SER, the relative mean is used to assess regional stability, as expressed by:
\begin{align}
	\label{eq8}
	r_l =\frac{m_l}{\max_{l} g -\min_{l} g}
\end{align}
where $m_l$ is the mean intensity of the DDS, $\max_{l} g$ and $\min_{l} g$ are the maximum and minimum intensities in the sequence $l$.
The termination constraint of this expansion is as follows:
\begin{equation}
\begin{aligned}
	\label{eq9}
	\Delta_{t}^l = \min(\frac{|\theta_{l}-\theta_{l-1}|}{th_\theta},\frac{|r_{l}-r_{l-1}|}{th_r})\\
\end{aligned}
\end{equation}
where $th_\theta$ retains its value in Eq. (\ref{eq6}), and $th_r$ is another threshold.

$\Delta_{t}^l$ is recalculated following each turn of expansion. If its value surpasses 1.0, this suggests that instability or partial offset has disrupted the condition of SER, leading to the termination of the expansion in this direction.
Once the expansions in both directions are finished, the area including the expanded and original DDSs is identified as a SER. 
Fig. \ref{fig9} shows the expansion process of the DDS marked by yellow in Fig. \ref{fig8}.

\subsection{Parameter estimation for CIS in SER}\label{sec4.3}
To achieve high-precision positioning, $g_a$ and $g_b$ can be acquired from the whole range of SER to acquire robust parameters.
We hope to estimate the stable bilateral strength by constructing a bilateral strength group and using the distribution parameters within the two populations and the relative stable strength difference between the two populations.
The calculation process of $g_a$ and $g_b$ on SER is explained below.

Firstly, pixels exhibiting the intensity that falls outside the mean plus variance are classified into high-intensity and low-intensity groups, denoted as $\mathbb{E}_a$ and $\mathbb{E}_b$, respectively. 
To obtain stable difference between the two groups, we calculate the difference between point pairs, which constructed by exhaustively combining each element from two groups. The value with the highest frequency is selected as the stable difference, denoted as $D_0$. 
Finally, according to the magnitude between the variance of $\mathbb{E}_a$ and $\mathbb{E}_b$, the evaluation of $g_a$ and $g_b$ is determined by the mean of these groups and $D_0$, the detail is shown in Algorithm 1. 

Fig. \ref{figm2} provides the positioning results of different methods. Both the CIS and fitting-based method exhibit a few outliers, as shown in Fig. \ref{figm2}(b) and (c). However, the incorporation of SER bolsters the robustness of positioning by eliminating these anomalous points and ensuring smoother edge transitions, as shown in Fig. \ref{figm2}(d).

\subsection{Edge complement based on Extension-Adjustment}\label{sec4.4}
\begin{figure}[H]
\centering
	\begin{minipage}[b]{.32\linewidth}
		\centering\subfigure[]{
			\includegraphics[width=0.5\linewidth]{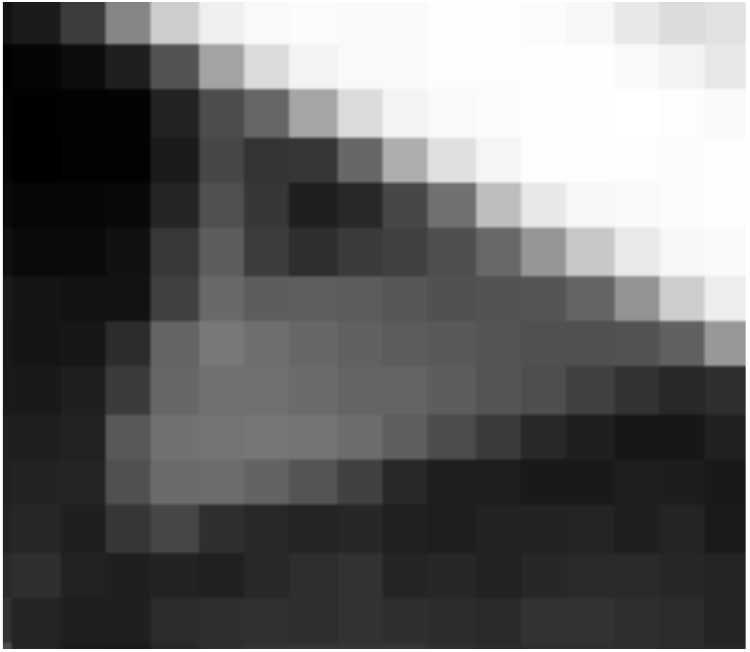}}
		\subfigure[]{
			\includegraphics[width=0.5\linewidth]{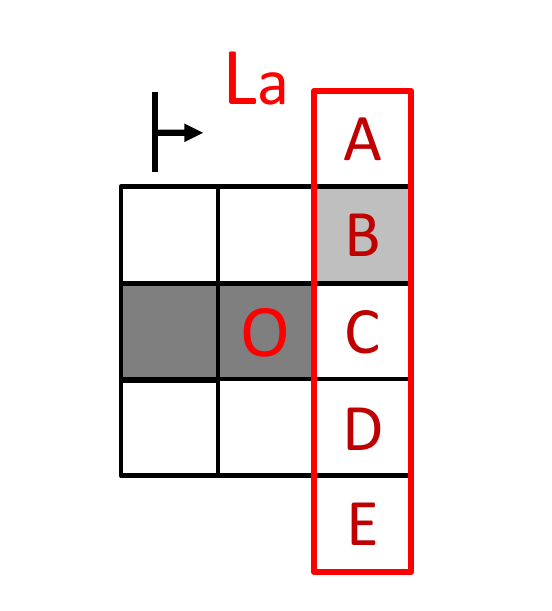}}
	\end{minipage} 
	\medskip
	\subfigure[]{
		\begin{minipage}[b]{.45\linewidth}
			\centering
			\includegraphics[width=0.85\linewidth]{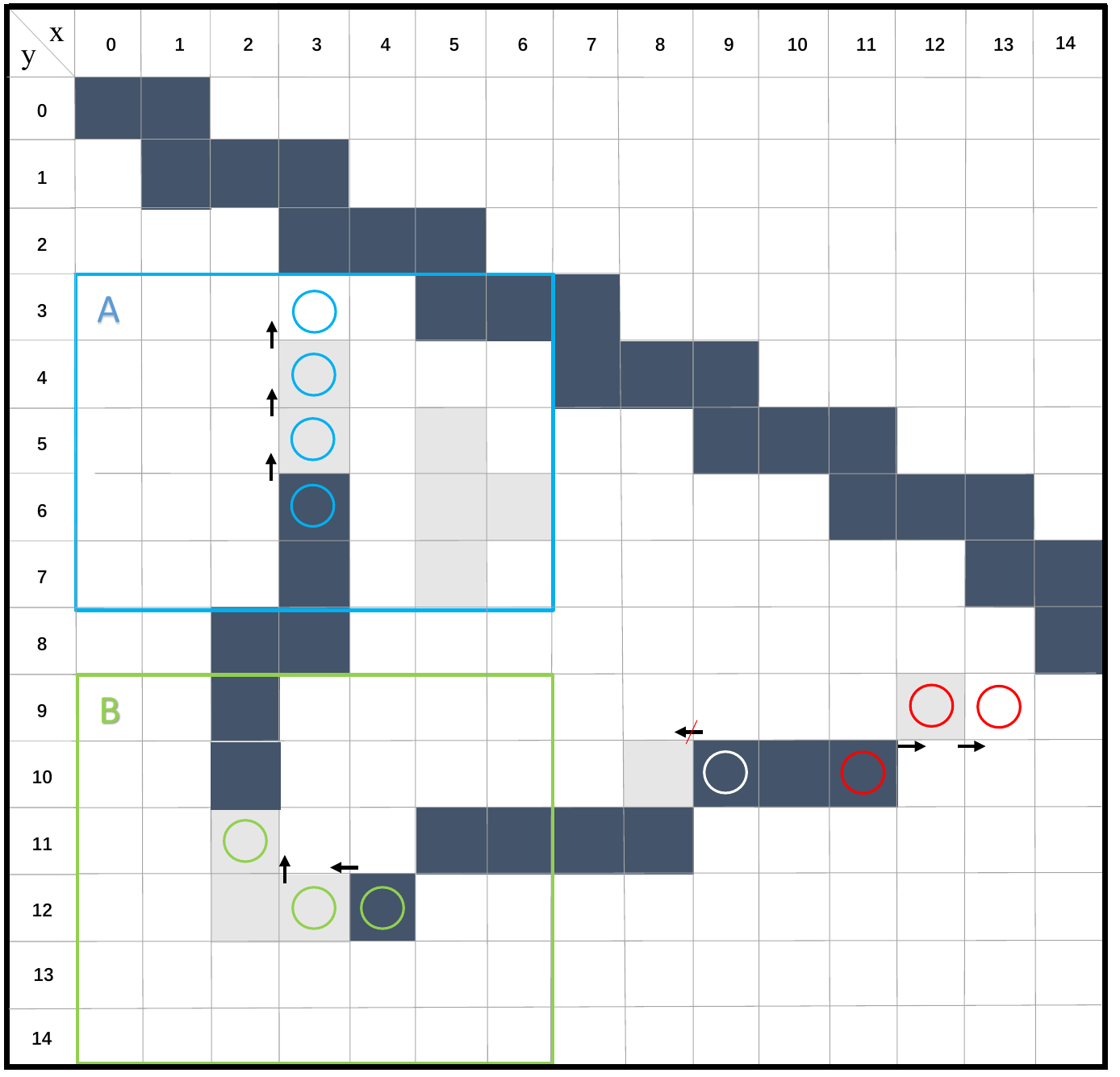}
	\end{minipage}}
	\caption{Conceptual diagram of the extension process. (a) the sub-image D form Fig. \ref{fig_s}. (b) The neighborhood of the pixel to be extended with the guide sequence $L_a$, which centres at (11, 10) in (c). (c) Extension process. Dark pixels $\in \mathbb{E}_S$ while gray pixels $\in \mathbb{E}_C$, and the circle marks the extended pixel.}
	\label{fig10}
\end{figure}

Due to the influence of various disturbances, not all edge pixels belong to stable regions. 
As illustrated in Fig. \ref{fig10}(c), for the set $\mathbb{E}$ of all edge pixels obtained by the edge detection algorithm during pre-processing, SER extracts stable pixels with locally regular intensity distribution, forming the SER set $\mathbb{E}_{S}$. 
This subsection proposes a new edge complement method based on an Extension-Adjustment strategy, which migrates SER to irregular regions. 
For the candidate set $\mathbb{E}_C=\{e_c | e_c\in \mathbb{E}, e_c\notin \mathbb{E}_{S}\}$, where $e_c$ is the candidate irregular edge pixel, the main steps are described as follows:

\textbf{Step 1}. Extension. If $e_c$ is situated adjacent to the endpoints of SERs, it will be selected and identified as the extended edge pixel. The extension process continues until no candidates presents nearby.

\textbf{Step 2}. Adjustment. The relevant parameters in SER are employed to adjust the irregular region, where the effective parameters of subpixel points are finally captured.

\subsubsection{Extension}
In Sec. \ref{sec4.2}, the extensible direction and range of angle are strictly restricted to guarantee the stability in SER. Given the complexity of edge directions in unstable areas, a guide sequence denoted by $L_a$ is constructed to determine the edge pixels to extend as well as the extension directions.
As shown in Fig. \ref{fig10}(b), when the extension direction is assumed to be on the right of the edge pixel $O$, $L_a$ is consists of five connected pixels from $A$ to $E$. And the possible pixels being extended, denoted as $B$, $C$, and $D$, which are located within the 8-neighbors of $O$ along the extension direction.

The following criterion is applied for extension process: as shown in Fig. \ref{fig10}(b), when both $A$ and $B$ are candidates $e_c$, $B$ is chosen as the extended pixel and the extending direction is going upwards. When $D$ and $E$ are candidates $e_c$, the extended pixel is then chosen as $D$ and extending direction is turning downwards. For all other cases, the extending direction remains unaltered and the edge extends to $C$. 

And the extension process is terminated when any point in $L_a$ does not belong to the set $\mathbb{E}$, or there are points from the set $\mathbb{E}_{S}$ present in the 8-neighbors of the current extended pixel.
A conceptual diagram illustrating the extension process is given in Fig. \ref{fig10}(c), which covers a variety of situations within the actual contexts.

\begin{figure}[H]
	\centering
	\subfigure[]{
		\includegraphics[width=1.5in]{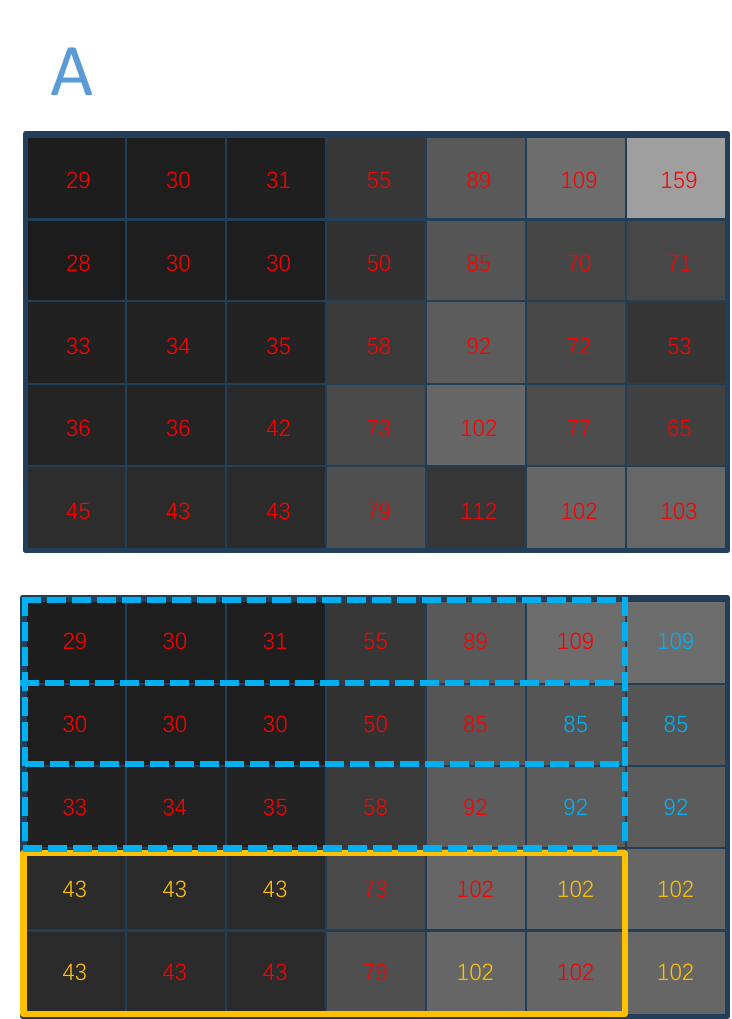} 
	}
	\hspace{0mm}
	\subfigure[]{
		\includegraphics[width=1.28in]{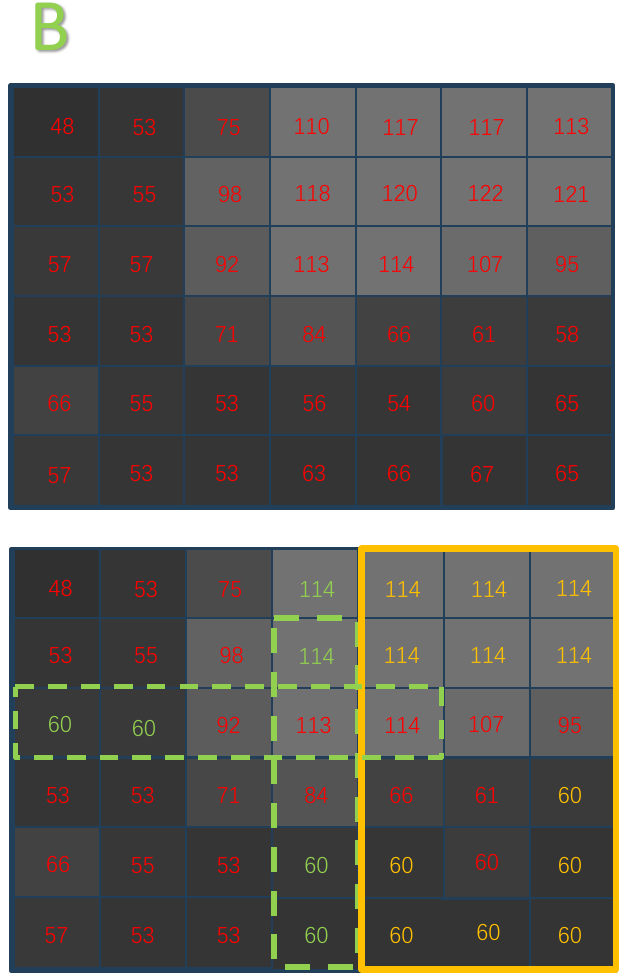} 
	}	
	\caption{Adjustment in different situations. The top row presents the original intensity distribution, which was derived from Fig. \ref{fig10}. The bottom row displays the adjusted intensity diagram, the red numbers denote the original intensity. The yellow region represents the SER after adjusting the intensity based on statistical parameters, while other colored numbers represent the adjusted results. (a) Local area $A$ which adjusts independently, where the dotted blue signifies the candidate sequence with adjustment. (b) Local area $B$ which the adjustment adheres to SER, where the dotted green line denotes the candidate sequence with adjustment.
	}
	\label{fig11}
\end {figure}

\subsubsection{Adjustment}

To realize the effective correction of irregular areas, it is necessary to extract essential information from the original SER, including the sequence length $L$ and the intensities $g^s_{a}$ and $g^s_{b}$ in two smooth resigns of SER.
Firstly, the candidate sequence of length $L$ is obtained, which centred on the extended pixel along the orientation of its guide sequence.
The intensities exhibiting the smallest difference between adjacent pixels are extracted from both sides of the sequence, which are labelled as $g^c_a$ and $g^c_b$.
Afterwards, the adjustment to intensity is performed, the criteria for judgement are as follows:
\begin{align}
	\label{eq10}
	D_c=\max (|g^s_{a}-g^c_a|,|g^s_{b}-g^c_b|)\\
	m_c= \frac{|(g^s_{a}+g^s_{b})-(g^c_a+g^c_b)|}{2}
\end{align}

When $m_c$ is below than a given threshold $th_c$ (defaults to 5), the candidate sequences are considered relatively stable, and adjustment strategies fall into the following two categories, as shown in Fig. \ref{fig11}. 
(a) Adjust independently. When $D_a$ exceeds $m_a$, the candidate sequence is viewed to be approximated as a stable sequence, with the intensities on both sides established as $g_l$ and $g_h$.
(b) Adjust according to SER. When $D_a$ is below or equal to $m_a$, the smooth areas of the candidate sequence are equivalent to the original SER and the intensities of the smooth sides are set to $g_{l0}$ and $g_{h0}$. 
In other cases, candidate sequences struggle to achieve a stable state and are subsequently discarded.
The judgment can be written as:
\begin{align}
	\begin{cases}
		g_a=g^c_a, g_b=g^c_b &\text{if } D_c > m_c, m_c\leq th_c\\
		g_a=g^s_a, g_b=g^s_b &\text{if } D_c \leq m_c, m_c\leq th_c\\
		\text{Discarded} &\text{if }m_c > th_c\\
	\end{cases}
\end{align}
Finally, subpixel edges in the adjusted candidate sequence are accurately localized by Eq. (\ref{eq4}).

\begin{figure}[t]
\centering
\begin{minipage}[b]{0.2\linewidth} 
    \centering
    \subfigure[]{
    \includegraphics[width=\textwidth]{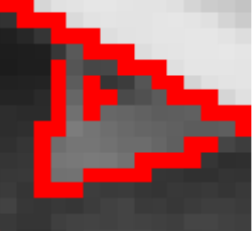}} 
    \end{minipage}
\hfill 
\begin{minipage}[b]{0.2\linewidth}
    \centering
    \subfigure[]{
    \includegraphics[width=\textwidth]{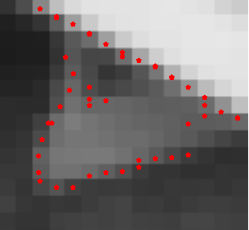}}
    \end{minipage}
\hfill
\begin{minipage}[b]{0.2\linewidth}
    \centering
    \subfigure[]{
    \includegraphics[width=\textwidth]{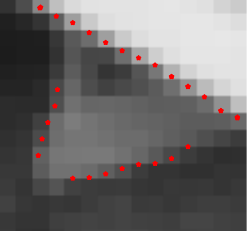}}
    \end{minipage}
\hfill
\begin{minipage}[b]{0.2\linewidth}
    \centering
    \subfigure[]{
    \includegraphics[width=\textwidth]{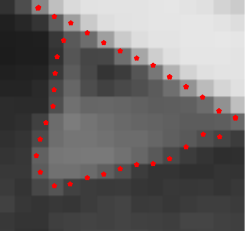}}
    \end{minipage}
\caption{The comparison of the detection results in the sub-image D form Fig. \ref{fig_s}. (a) Pixel-level edge detection by Canny. (b) Sub-pixel edge localization by CIS, where large levels of errors exist in tricky context. (c) Sub-pixel edge localization by CIS + SER without the edge complement, which removes outliers in the interference area. (d) Sub-pixel edge localization by CIS + SER including the edge complement, which pads the discontinuous edges.}
\label{figm3}
\end{figure}

Fig. \ref{figm3} presents the detection results under various strategies for our method. As shown in Fig. \ref{figm3}(b), CIS can effectively locate simple edges, but it exhibits numerous errors at intersections or disturbances due to its strict reliance on edge detection (as shown in Fig. \ref{figm3}(a)). The introduction of SER eliminates both interfering points and outliers, but leading to the discontinuous edges, as seen in Fig. \ref{figm3}(c). Nevertheless, these gaps are successfully bridged by implementing edge complement, producing a more complete and coherent edge, as shown in Fig. \ref{figm3}(d). It is observed that the overall algorithm proposed can effectively remove the interference or invalid edges, while maintaining exceptional robustness in scenes involving crossed edges. Ultimately, it achieves the coherent and clear sub-pixel edge positioning.

\begin{figure}[b]
		\centering
		\subfigure[]
		{
			\begin{minipage}[b]{.18\linewidth}
				\centering
				\includegraphics[width=0.7in]{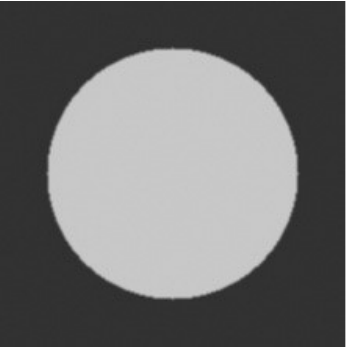} 	 \\
				\vspace{0.2mm}
				\includegraphics[width=0.7in]{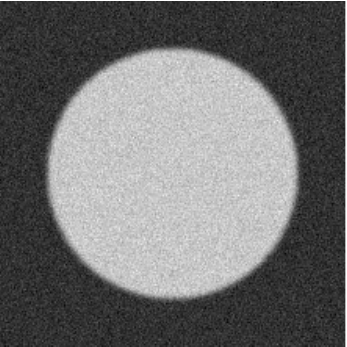} 	
			\end{minipage}
		}
		\subfigure[]
		{
			\begin{minipage}[b]{.3\linewidth}
				\centering
				\includegraphics[width=1.2in]{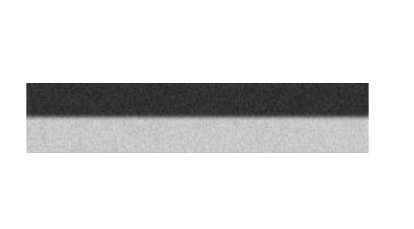} 	 \\
				\vspace{0.2mm}
				\includegraphics[width=1.2in]{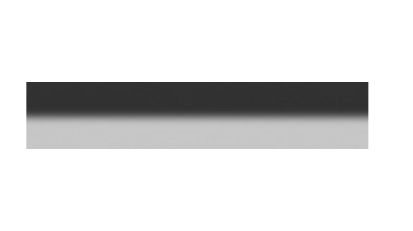} 	
			\end{minipage}
		}
		\subfigure[]
		{
			\begin{minipage}[b]{.18\linewidth}
				\centering
				\includegraphics[width=.7in]{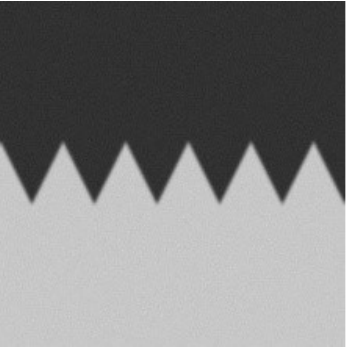} 	\\
				\vspace{0.2mm}
				\includegraphics[width=.7in]{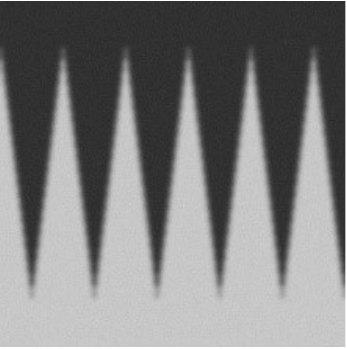} 	
			\end{minipage}
		}
		\caption{Examples of three simulated datasets. (a) The sample of circle dataset. The top one is with the parameter setting as $k_G=3$ and $SNR =100$; the bottom one is with $k_G=9$ and $SNR =70$.
			(b) The sample of line dataset. The top one is with the parameter setting as $\sigma_L=1$, $SNR=70$, and $L=-0.3$; the bottom one is with $\sigma_L=2.25$, $SNR=85$, and $L=0.3$.
			(c) The sample of slant dataset. The top one is with parameter setting as $k_G=5$, $SNR=85$, and slope $=$ 2; the bottom one is with $k_G=7$, $SNR=85$, and slope $=$ 8.}
		\label{fig12}
	\end{figure}
	
\section{Experimental results and analysis}\label{sec5}
In this section, a comprehensive series of experiments are conducted to evaluate the efficacy of our method. 
All computations are implemented with python 3.7 on a computer with AMD Ryzen 7 4800H CPU GHz and 4GB RAM.

\begin{figure}[b]
\centering
\includegraphics[width=2.5in]{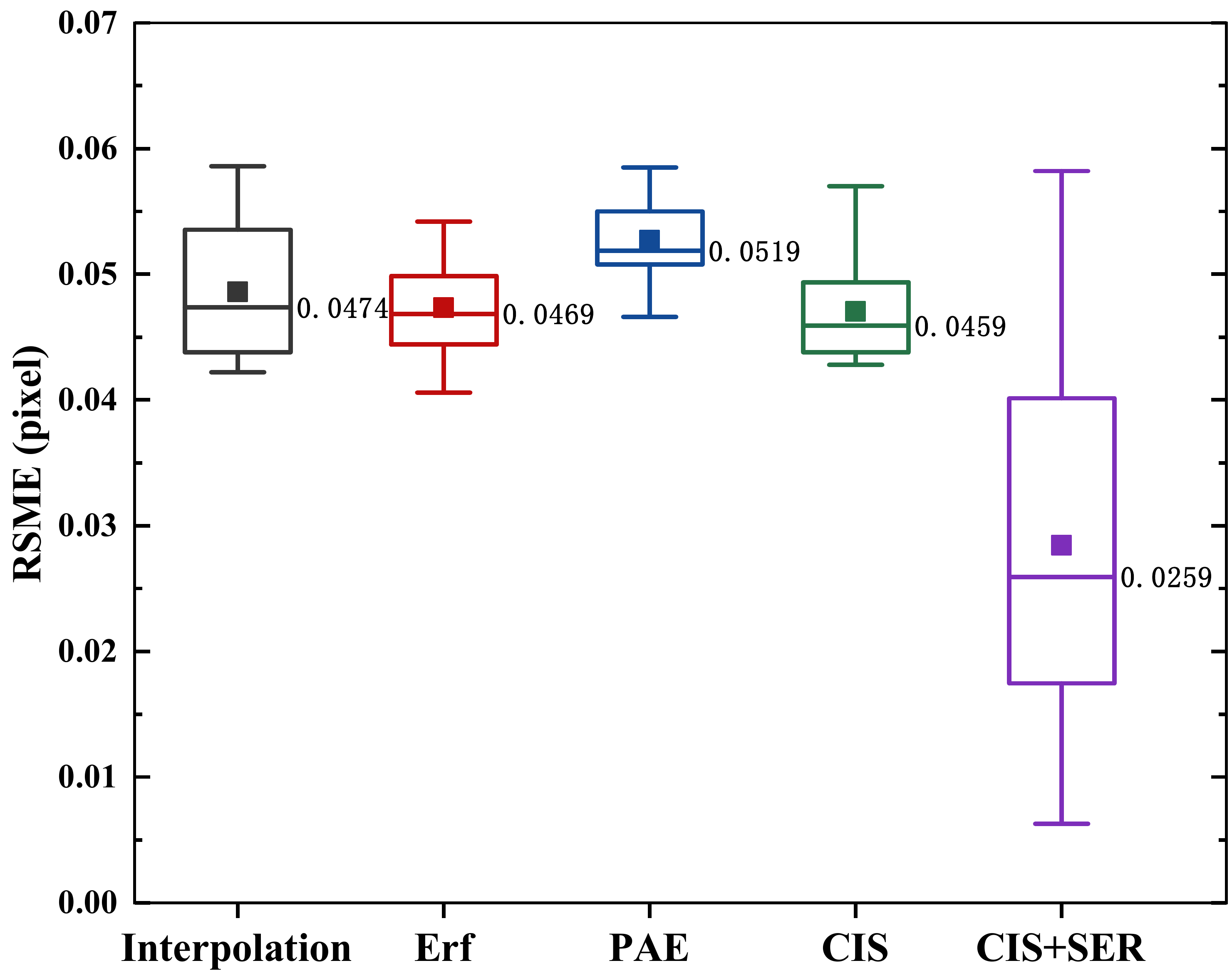}
\caption{Box plot of the results on circle dataset. Data are taken from Table \ref{tab2} except for $SNR=70$, where the numbers represent the median values.} \label{fig13}
\end{figure}

\subsection{Experiments on simulated images}\label{sec5.1}
To evaluate the accuracy and robustness of the proposed method under complex conditions, we provides the quantitative analysis in a series of simulated datasets, comprising circle, line, and slant, as shown in Fig. \ref{fig12}. 
The first two datasets are employed to examine the accuracy and robustness against noise and blurring. The slant dataset is specifically designed to assess the impact of mutual interference between edges. Details regarding these datasets will be described later.

The edges with different blurred levels are generated by Gaussian filters with various size ($k_G$ * $k_G$). And the intensity of Gaussian noise is determined by signal-to-noise ratio ($SNR$), defined as
\begin{equation}
SNR=20\log_{10}\frac {k_n}{\sigma_n}
\end{equation}
where $k_n$ is the intensity difference between foreground and background, while $\sigma_n$ is the standard deviation of the noise. All sample images in these datasets are generated by combinations of various edge patterns and noise levels.

Several classical and state-of-the-art methods are used for quantative comparison in the subpixel edge positioning experiments, including Zernike moment\cite{Ghosal1993OrthogonalMO}, polynomial interpolation \cite{Qingli2003AIS}, fitting based on Erf\cite{2011Edge}, the partial area effect (PAE)\cite{TrujilloPino2013AccurateSE}, Canny/Devernay\cite{Gioi2017ASE}, and AWG\cite{Seo2018SubpixelEL}. Additionally, the CIS described in the Sec. \ref{sec3} also participates the comparison, so does the compound method, referred to as CIS+SER, which covers the whole framework. The primary parameters of the part algorithm are presented in Table \ref{tabc}.

\begin{table}[t!]
	\centering
	\begin{threeparttable}
	\newcommand{\tabincell}[2]{\begin{tabular}{@{}#1@{}}#2\end{tabular}}
		\caption{The statistical results for verification}
		\label{tabc}
		\renewcommand\arraystretch{1.3}
		\begin{tabular}{cc}
			\hline
			Method & Parameters\\ 
			\hline
			Canny & $th_l=80$, $th_h=100$ \\
			Erf \& CIS \& interpolation & $n_p=7$ \\
			Zernike moment & \tabincell{c}{$ifmedianBlur=true$, $nbsize=7$,\\ $th_k=20.0, th_l=\sqrt{2}/nbsize$}\\ 
			PAE  &  $th=10$, $iters =0$, $order=2$\\ 
			Canny/Devernay & $\sigma=3.0$, $th_l=2.0$, $th_h=2.9$ \\ 
			CIS + SER &   \tabincell{c}{$n_p=7$, $th_m=5$, $th_\theta=\pi/40$, \\ $th_{ea}=10$, $th_r=0.1$, $th_c=10$}\\  
			\hline
		\end{tabular}
	\end{threeparttable}
\end{table}

\begin{table}[t]
\centering
\begin{threeparttable}
\newcommand{\tabincell}[2]{\begin{tabular}{@{}#1@{}}#2\end{tabular}}
\caption{Errors of fitting radius in the circle dataset.}
\label{tab2}
\renewcommand\arraystretch{1.2}
   \begin{tabular}{ccccccccc}
\hline
\multicolumn{1}{c}{$k_G$} & \multicolumn{1}{c}{$SNR$} & Zernike &\tabincell{c}{Inter \\ polation}  & Erf & PAE &  \tabincell{c}{Canny / \\ Devernay} & CIS & \tabincell{c}{CIS +\\SER} \\ 
\hline
\multicolumn{1}{c}{\multirow{3}{*}{3}} & \multicolumn{1}{c}{80}  & 1.4197  & 0.0430  & \textit{0.0406}  & 0.0466 & 0.1118 & 0.0437 & \textit{\textbf{0.0280}} \\
\multicolumn{1}{c}{}  & \multicolumn{1}{c}{90}  & 1.4616  & \textit{0.0422} & 0.0432 & 0.0529 & 0.0927 & 0.0428 & \textit{\textbf{0.0390}} \\
\multicolumn{1}{c}{}  & \multicolumn{1}{c}{100} & 1.4787  & \textit{0.0423} & 0.0451 & 0.0513 & 0.1031 & 0.0432 & \textit{\textbf{0.0413}}  \\ 
\hline
\multicolumn{1}{c}{\multirow{3}{*}{5}}  & \multicolumn{1}{c}{80}  & 1.5356  & 0.0462 & 0.0469 & 0.0516 & 0.1217 & \textit{0.0443}  & \textit{\textbf{0.0132}}  \\
\multicolumn{1}{c}{} & \multicolumn{1}{c}{90}  &1.5492  & 0.0446  & \textit{0.0437} & 0.0521 & 0.0936 & 0.0439 & \textit{\textbf{0.0156}} \\
\multicolumn{1}{c}{}  & \multicolumn{1}{c}{100} & 1.5746  & \textit{0.0448} & 0.0466 & 0.0532 & 0.1131 & 0.0470 & \textit{\textbf{0.0063}} \\ 
\hline
\multicolumn{1}{c}{\multirow{3}{*}{7}} & \multicolumn{1}{c}{80}  & 1.7526  & 0.0544  & \textit{0.0472}  & 0.0585 & 0.1110  & 0.0507 & \textit{\textbf{0.0193}}  \\
\multicolumn{1}{c}{}  & \multicolumn{1}{c}{90} & 1.9076  & 0.0489  & 0.0468 & 0.0516 & 0.1321 & \textit{0.0448} & \textit{\textbf{0.0262}}\\
\multicolumn{1}{c}{}  & \multicolumn{1}{c}{100} & 1.9607  & 0.0485  & 0.0482 & 0.0503 & 0.1284 & \textit{0.0480} & \textit{\textbf{0.0237}} \\
\hline
\multicolumn{1}{c}{\multirow{3}{*}{9}} & \multicolumn{1}{c}{80}  & 1.9862  & 0.0568  & \textit{0.0515}  & 0.0568 & 0.0962 & 0.0570 & \textit{\textbf{0.0256}} \\
\multicolumn{1}{c}{}  & \multicolumn{1}{c}{90}  & 2.1611  & 0.0586  & 0.0542 & 0.0585 & 0.1041 & \textit{0.0514}  & \textit{\textbf{0.0447}}\\
\multicolumn{1}{c}{}  & \multicolumn{1}{c}{100}  & 2.1830 & 0.0527  & 0.0539 & \textit{0.0489} & 0.0994 & \textit{\textbf{0.0477}} & 0.0582 \\
\hline
\multicolumn{2}{c}{Mean} & 1.7476  & 0.0486 & 0.0474 & 0.0527 & 0.1089 & \textit{0.0466}  & \textit{\textbf{0.0284}} \\ 
\hline
    \end{tabular}
    \begin{tablenotes}[para,flushleft]
  \item \textit{\textbf{Bold italic number}}: the minimum. \textit{Italic number}: the second smallest.
    \end{tablenotes}
\end{threeparttable}
\end{table}

\begin{table}[t]
\centering
\begin{threeparttable}
\newcommand{\tabincell}[2]{\begin{tabular}{@{}#1@{}}#2\end{tabular}}
\caption{Errors of fitting radius in the circle dataset.}
\label{tab2}
\renewcommand\arraystretch{1.2}
   \begin{tabular}{ccccccccc}
\hline
\multicolumn{1}{c}{$k_G$} & \multicolumn{1}{c}{$SNR$} & Zernike &\tabincell{c}{Inter \\ polation}  & Erf & PAE &  \tabincell{c}{Canny / \\ Devernay} & CIS & \tabincell{c}{CIS +\\SER} \\ 
\hline
\multicolumn{1}{c}{\multirow{4}{*}{3}} & \multicolumn{1}{c}{70} & 1.5354  & 2.6864 & 2.3053 & 1.8185  & \textit{\textbf{0.4005}} & 2.3333 & \textit{0.5278}  \\
\multicolumn{1}{c}{}  & \multicolumn{1}{c}{80}  & 1.4197  & 0.0430  & \textit{0.0406}  & 0.0466 & 0.1118 & 0.0437 & \textit{\textbf{0.0280}} \\
\multicolumn{1}{c}{}  & \multicolumn{1}{c}{90}  & 1.4616  & \textit{0.0422} & 0.0432 & 0.0529 & 0.0927 & 0.0428 & \textit{\textbf{0.0390}} \\
\multicolumn{1}{c}{}  & \multicolumn{1}{c}{100} & 1.4787  & \textit{0.0423} & 0.0451 & 0.0513 & 0.1031 & 0.0432 & \textit{\textbf{0.0413}}  \\ 
\hline
\multicolumn{1}{c}{\multirow{4}{*}{5}}  & \multicolumn{1}{c}{70}  & 1.5339  & 3.0131 & 2.6623 & 2.3043 & \textit{1.0204} & 2.7992 & \textit{\textbf{0.1846}} \\
\multicolumn{1}{c}{}  & \multicolumn{1}{c}{80}  & 1.5356  & 0.0462 & 0.0469 & 0.0516 & 0.1217 & \textit{0.0443}  & \textit{\textbf{0.0132}}  \\
\multicolumn{1}{c}{} & \multicolumn{1}{c}{90}  &1.5492  & 0.0446  & \textit{0.0437} & 0.0521 & 0.0936 & 0.0439 & \textit{\textbf{0.0156}} \\
\multicolumn{1}{c}{}  & \multicolumn{1}{c}{100} & 1.5746  & \textit{0.0448} & 0.0466 & 0.0532 & 0.1131 & 0.0470 & \textit{\textbf{0.0063}} \\ 
\hline
\multicolumn{1}{c}{\multirow{4}{*}{7}} & \multicolumn{1}{c}{70}  & 1.6705  & 3.0398  & 2.3996 & 2.0371 & \textit{1.6516} & 2.7415 & \textit{\textbf{0.2800}} \\
\multicolumn{1}{c}{}  & \multicolumn{1}{c}{80}  & 1.7526  & 0.0544  & \textit{0.0472}  & 0.0585 & 0.1110  & 0.0507 & \textit{\textbf{0.0193}}  \\
\multicolumn{1}{c}{}  & \multicolumn{1}{c}{90} & 1.9076  & 0.0489  & 0.0468 & 0.0516 & 0.1321 & \textit{0.0448} & \textit{\textbf{0.0262}}\\
\multicolumn{1}{c}{}  & \multicolumn{1}{c}{100} & 1.9607  & 0.0485  & 0.0482 & 0.0503 & 0.1284 & \textit{0.0480} & \textit{\textbf{0.0237}} \\
\hline
\multicolumn{1}{c}{\multirow{4}{*}{9}} & \multicolumn{1}{c}{70} & 1.8172 & 3.5815 & 2.3905 & 3.5174 & \textit{1.1867}  & 3.0019 & \textit{\textbf{0.7520}}  \\ 
\multicolumn{1}{c}{}& \multicolumn{1}{c}{80}  & 1.9862  & 0.0568  & \textit{0.0515}  & 0.0568 & 0.0962 & 0.0570 & \textit{\textbf{0.0256}} \\
\multicolumn{1}{c}{}  & \multicolumn{1}{c}{90}  & 2.1611  & 0.0586  & 0.0542 & 0.0585 & 0.1041 & \textit{0.0514}  & \textit{\textbf{0.0447}}\\
\multicolumn{1}{c}{}  & \multicolumn{1}{c}{100}  & 2.1830 & 0.0527  & 0.0539 & \textit{0.0489} & 0.0994 & \textit{\textbf{0.0477}} & 0.0582 \\
\hline
\multicolumn{2}{c}{Mean} & 1.7205  & 0.8065 & 0.6453 & 0.6443 & \textit{0.3479} & 0.7150  & \textit{\textbf{0.1303}} \\ 
\hline
    \end{tabular}
    \begin{tablenotes}[para,flushleft]
  \item \textit{\textbf{Bold italic number}}: the minimum. \textit{Italic number}: the second smallest.
    \end{tablenotes}
\end{threeparttable}
\end{table}

\subsubsection{Analysis of circle dataset}
The size of image is 221 × 221, and the circle with radius 80 is in the center of the image. The internal intensity is 200 while the external intensity is 50. The size of Gaussian filters $k_G$ is set to four levels: 3, 5, 7, and 9. $SNR$ are increased from 70 to 100 dB with 10 dB steps. 
Each subset covering the mentioned factors contains 5 samples. The errors are defined as the difference between the actual radius and the fitting radius, which is derived from the obtained subpixel edge coordinates and the actual center.
\begin{figure*}[b!]
\centering
    \includegraphics[scale=0.24]{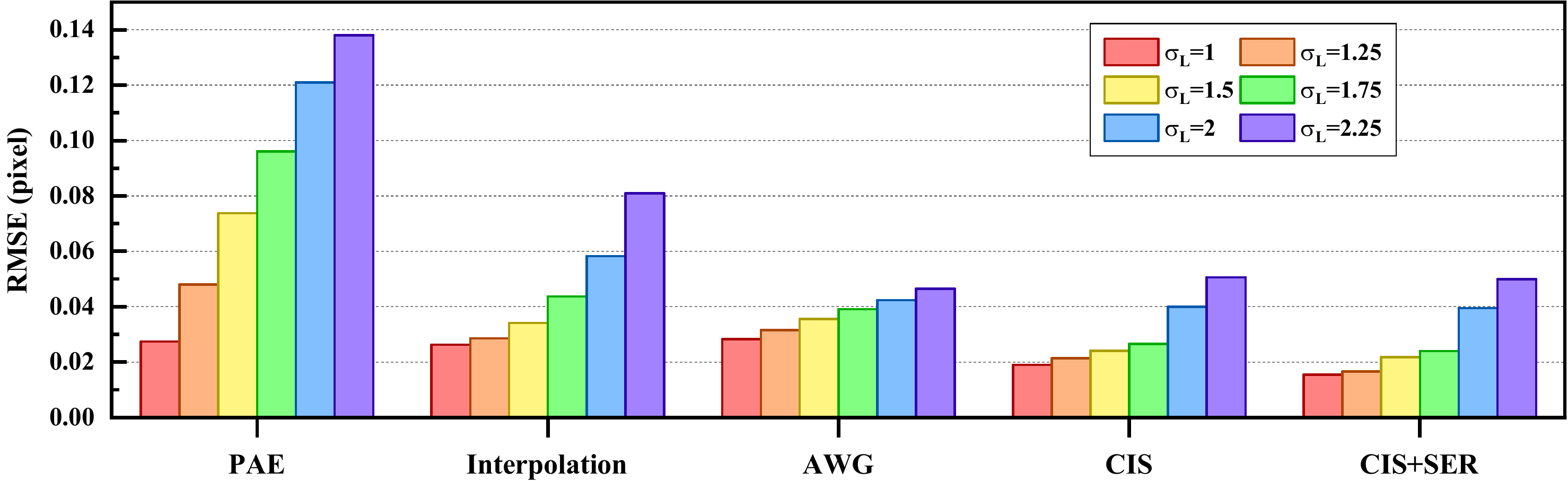}
\caption{Mean RMSE of each method under different $\sigma_L$ on line dataset.} 
\label{fig14}
\end{figure*}

\begin{table}[t]
\centering
\begin{threeparttable} 
\newcommand{\tabincell}[2]{\begin{tabular}{@{}#1@{}}#2\end{tabular}}
\caption{rmse in the line dataset.}
\label{tab3}
\renewcommand\arraystretch{1.2}
\begin{tabular}{ccccccc}
\hline
\multicolumn{1}{c}{$\sigma_L$} & \multicolumn{1}{c}{$SNR$} & PAE & \tabincell{c}{Inter polation}  & AWG & CIS & \tabincell{c}{CIS+SER} \\ \hline
\multirow{5}{*}{1.00} & 85  & 0.0274 & 0.0257  & \textit{\textbf{0.0094}} & 0.0183  & \textit{0.0156}  \\
 & 79  & 0.0275 & 0.0269  & 0.0189  & \textit{0.0186}  & \textit{\textbf{0.0148}} \\
 & 76  & 0.0273 & 0.0271  & 0.0289  & \textit{0.0194}  & \textit{\textbf{0.0159}} \\
 & 73  & 0.0272 & 0.0258  & 0.0378  & \textit{0.0189}  & \textit{\textbf{0.0149}} \\
 & 70  & 0.0276 & 0.0258  & 0.0462  & \textit{0.0198}  & \textit{\textbf{0.0162}} \\ \hline
\multicolumn{2}{c}{Mean}   & 0.0274 & 0.0263  & 0.0282  & \textit{0.0190}  & \textit{\textbf{0.0155}} \\ \hline
\multirow{5}{*}{1.25} & 85  & 0.0480 & 0.0285  & \textit{\textbf{0.0105}} & 0.0213  & \textit{0.0162} \\
 & 79  & 0.0481 & 0.0293  & 0.0211  & \textit{0.0207}  & \textit{\textbf{0.0159}} \\
 & 76  & 0.0480 & 0.0276  & 0.0314  & \textit{0.0218}  & \textit{\textbf{0.0167}} \\
 & 73  & 0.0481 & 0.0280  & 0.0427  & \textit{0.0209}  & \textit{\textbf{0.0169}} \\
 & 70  & 0.0480 & 0.0294  & 0.0522  & \textit{0.0221}  & \textit{\textbf{0.0172}} \\ \hline
\multicolumn{2}{c}{Mean}   & 0.0480 & 0.0286  & 0.0316  & \textit{0.0214}  & \textit{\textbf{0.0166}} \\ \hline
\multirow{5}{*}{1.50}  & 85  & 0.0734 & 0.0342  & \textit{\textbf{0.0117}} & 0.0243  & \textit{0.0200}  \\
 & 79  & 0.0742 & 0.0334  & \textit{0.0235}  & \textit{0.0240}  & \textit{\textbf{0.0221}} \\
 & 76  & 0.0739 & 0.0349  & 0.0354  & \textit{0.0241}  & \textit{\textbf{0.0222}} \\
 & 73  & 0.0737 & 0.0341  & 0.0474  & \textit{0.0236}  & \textit{\textbf{0.0220}} \\
 & 70  & 0.0735 & \textit{0.0340} & 0.0596  & \textit{0.0244}  & \textit{\textbf{0.0228}} \\ \hline
\multicolumn{2}{c}{Mean}   & 0.0738 & 0.0341  & 0.0355  & \textit{0.0241} & \textit{\textbf{0.0218}} \\ \hline
\multirow{5}{*}{1.75} & 85  & 0.0961 & 0.0438  & \textit{\textbf{0.0130}} & 0.0264  & \textit{0.0239} \\
 & 79  & 0.0959 & \textit{0.0422} & \textit{0.0257}  & 0.0266  & \textit{\textbf{0.0239}} \\
 & 76  & 0.0961 & \textit{0.0413} & 0.0392  & \textit{0.0269}  & \textit{\textbf{0.0243}} \\
 & 73  & 0.0967 & 0.0443  & 0.0517  & \textit{0.0270}  & \textit{\textbf{0.0230}} \\
 & 70  & 0.0957 & 0.0467  & 0.0661  & \textit{0.0261}  & \textit{\textbf{0.0250}} \\ \hline
\multicolumn{2}{c}{Mean}   & 0.0961 & 0.0437  & 0.0391  & \textit{0.0266}  & \textit{\textbf{0.0240}} \\ \hline
\multirow{5}{*}{2.00} & 85  & 0.1208 & 0.0560  & \textit{\textbf{0.0140}} & 0.0408  & \textit{0.0397}  \\
 & 79  & 0.1213 & 0.0579  & \textit{\textbf{0.0286}} & 0.0402  & \textit{0.0397}  \\
 & 76  & 0.1205 & 0.0597  & 0.0431  & \textit{0.0382}  & \textit{\textbf{0.0375}} \\
 & 73  & 0.1213 & 0.0585  & 0.0536  & \textit{\textbf{0.0395}} & \textit{0.0403}  \\
 & 70  & 0.1212 & 0.0593  & 0.0726  & \textit{0.0412}  & \textit{\textbf{0.0402}} \\ \hline
\multicolumn{2}{c}{Mean}   & 0.1210 & 0.0583  & 0.0424  & \textit{0.0400}  & \textit{\textbf{0.0395}} \\ \hline
\multirow{5}{*}{2.25} & 85 & 0.1380 & 0.0749  & \textit{\textbf{0.0152}} & 0.0510  & \textit{0.0501}  \\
 & 79  & 0.1380 & 0.0838  & \textit{\textbf{0.0306}} & 0.0505  & \textit{0.0499}  \\
 & 76  & 0.1381 & 0.0779  & \textit{\textbf{0.0458}} & 0.0505  & \textit{0.0493}  \\
 & 73  & 0.1375 & 0.0806  & 0.0610  & \textit{\textbf{0.0505}} & \textit{0.0505}  \\
 & 70  & 0.1381 & 0.0874  & 0.0798  & \textit{0.0505}  & \textit{\textbf{0.0498}} \\ \hline
\multicolumn{2}{c}{Mean} & 0.1379 & 0.0809  & \textit{\textbf{0.0465}} & 0.0506  & \textit{0.0499}  \\ \hline
\end{tabular}
\begin{tablenotes}[para,flushleft]
  \item \textit{\textbf{Bold italic number}}: the minimum. \textit{Italic number}: the second smallest.
\end{tablenotes}
\end{threeparttable}
\end{table}

Table \ref{tab2} illustrates the performance of different methods on the circle dataset. CIS+SER exhibits the smallest deviation among all methods, and notably, CIS also achieves competitive results and maintains suboptimal positioning accuracy in most cases.
When $SNR$ equal to 70, the accuracy of other methods significantly decreases, but CIS+SER can sustain a relatively minor error, demonstrating the validity of the stable region. 

Excluding Zernike and Canny/Devernay, the methods consistently maintain errors below 0.06 when $SNR \geq 80$, the results are then illustrated in a box plot, as depicted in Fig. \ref{fig13}.
Among the above methods, PAE proves to be the most robust. However, its median error is limited to 0.0519 pixels, while interpolation, Erf, and CIS exhibit median errors of around 0.047 pixels.
And the median error of CIS+SER is further reduced to 0.0259 pixels.
Nevertheless, this enhancement is accompanied by a higher variance, signifying uncertainty regarding its actual impact. Given that the method depends on the intensity distribution on the edge region, higher errors are observed at low noises ($SNR\geq90$) when $k_G$ = 3 or 9.
However, it is evident that SER can exert a beneficial influence on CIS in most instances.

\begin{figure}[t]
\centering
    \subfigure[]{
    \includegraphics[width=2in]{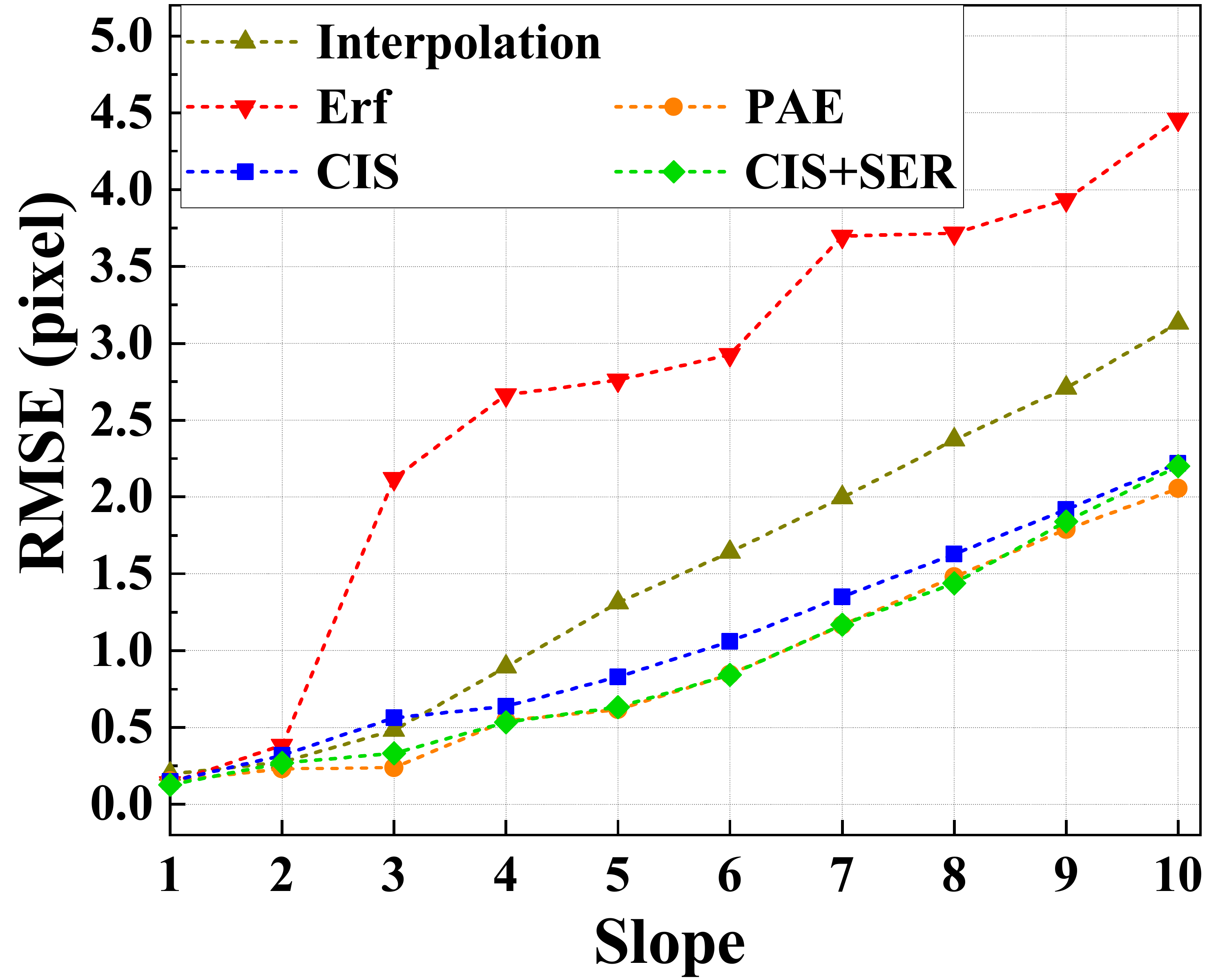} 
    }
    \subfigure[]{
    \includegraphics[width=2in]{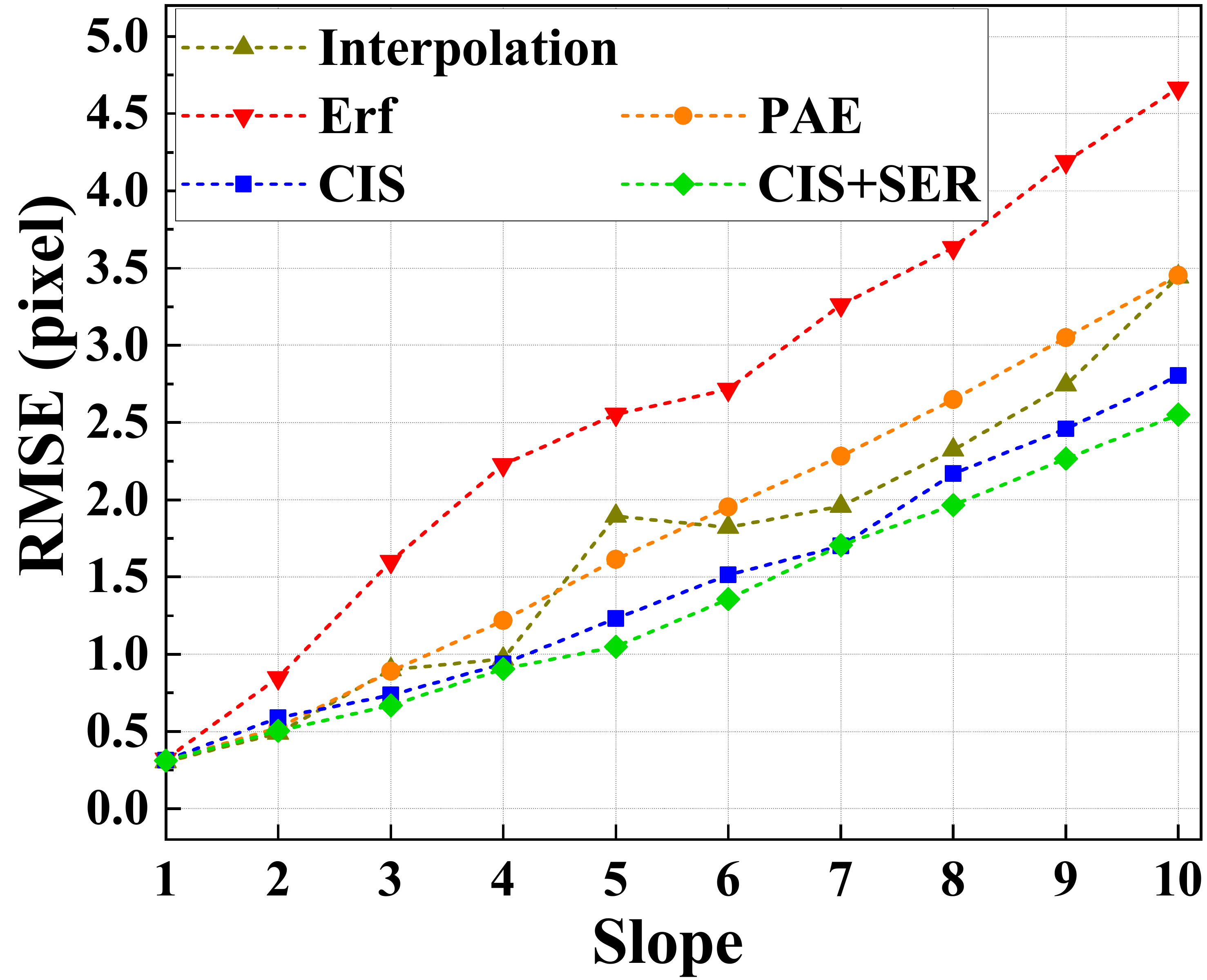} 
    }	
    \caption{Comparison of the subpixel edge positioning accuracy of the five
methods on the slant dataset. The slope was set to vary from 1 to 10 and $SNR=85$ for all data. (a) $k_G=5.$ (b) $k_G=7.$}
\label{fig15}
\end{figure}

\subsubsection{Analysis of line dataset}

Referring to the design in \cite{Seo2018SubpixelEL}, edges in the line dataset are generated by the following equation:
\begin{align}
f_L(x)=\frac{D_L}{2}\left ( erf(\frac{x-L}{\sqrt{2} \sigma _L} ) + 1 \right ) + I_L
\end{align}
where $f_L(x)$ is the intensity of edge point $x$, $L$ is the edge location, and $\sigma _L$ is the blur coefficient. $I_L$ is the intensity in one side, and $D_L$ is the difference in intensity between the two sides. In our experiment, $I_L$ = $50$ and $D_L$ = $150$ .

The size of the sample is 200 × 40, the edge in the image is generated by a combination of varying blur coefficient values, noise levels, and edge locations. 
$SNR$ is set to the following values respectively: 70, 73, 76, 79, 85, and $\sigma_L$ is set in the range from 1 to 2.25 with 0.25 intervals. 
The edge location $L$ is situated at the vertical center in the image, with a deviation from -0.5 to 0.5 pixels at an increment of 0.1 pixels.
Samples with identical $\sigma_L$ and $SNR$ values are viewed as a group, and the the root mean square error (RMSE) is calculated between the real and measured coordinates of $L$.

As shown in Table \ref{tab3}, our method obtains optimal or sub-optimal accuracy in comparison with other methods. Although AWG achieves the minimum positioning error when $SNR$ = $85$, it is more susceptible to noise interference. Specifically, while our method is suboptimal under low noise, it achieves the lowest mean errors under other noise levels due to its robust anti-interference capability.

Overall, the mean accuracy of these methods is less influenced by $SNR$ and more affected by $\sigma_L$. Fig. \ref{fig14} shows the mean error under different $\sigma_L$. Our methods maintain errors at around 0.02 pixels when $\sigma_L$ is below 2. Moreover, with the incorporation of SER further enhances CIS, its accuracy surpassing that of other methods when $\sigma_L$ values are 1 and 1.25.
It is worth noting that the accuracy improvement on SER for CIS is not as significant as the result in the circular dataset. This is because the edge in lines are inherently less complex than circles, allowing CIS to achieve relatively optimal measurements without incorporating SER.

\begin{figure}[b!]
\centering
    \subfigure[]{
	\includegraphics[width=1in]{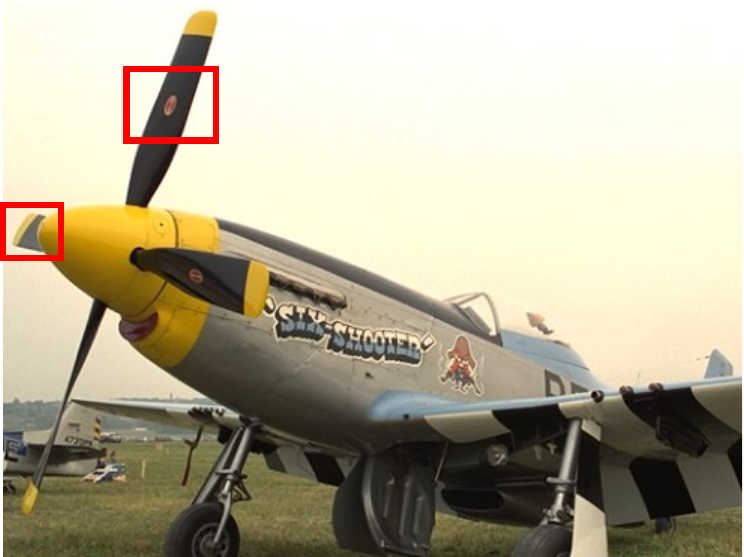} 
    }
	\subfigure[]{
	\includegraphics[width=1in]{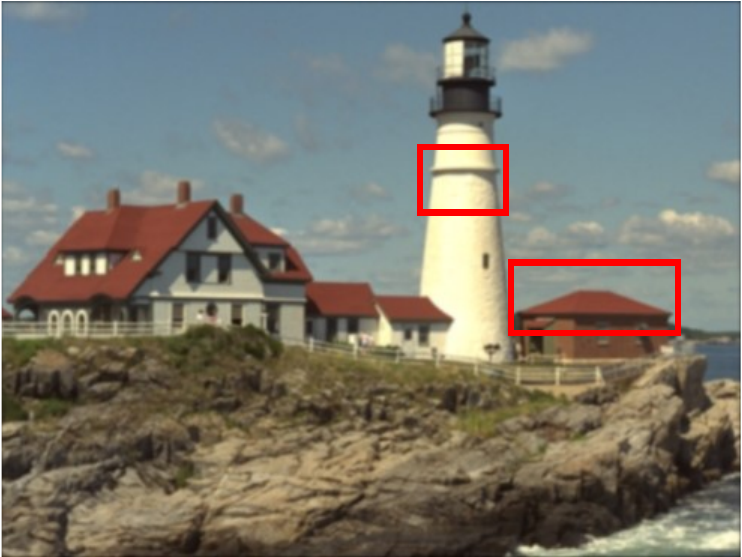} 
    }	
    \subfigure[]{
	\includegraphics[width=1in]{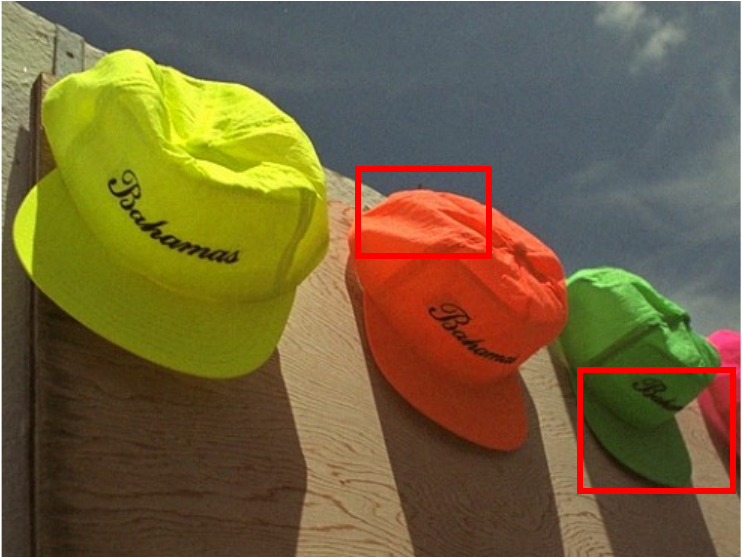} 
    }	
	\caption{Images from the TID2013. The sub-images bounded by the red rectangles are used for testing in Fig. \ref{fig17}. (a) Image without any interferences. (b) Image with Gaussian noise added. (c) Image smoothed by Gaussian filter.}
    \label{fig16}
\end {figure}

\begin{figure*}[t!]
\centering
\subfigure
{\rotatebox{90}{\scriptsize{~~~~~~~~~~\rotatebox{-90}{\textbf{Image 6}}~~~~~~~~~~~~~~~\rotatebox{-90}{\textbf{Image 5}}~~~~~~~~~~~~~~\rotatebox{-90}{\textbf{Image 4}}~~~~~~~~~~~~~
\rotatebox{-90}{\textbf{Image 3}}~~~~~~~~~~~~~~~~~~\rotatebox{-90}{\textbf{Image 2}}~~~~~~~~~~~~~~~~~~~~~\rotatebox{-90}{\textbf{Image 1}}}}}
\setcounter{subfigure}{0}
\subfigure[]
{
  \begin{minipage}[b]{.11\linewidth}
  \centering
  \includegraphics[width=1.15\textwidth]{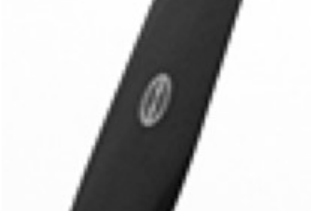} \\\vspace {0.5mm}
  \includegraphics[width=1.15\textwidth]{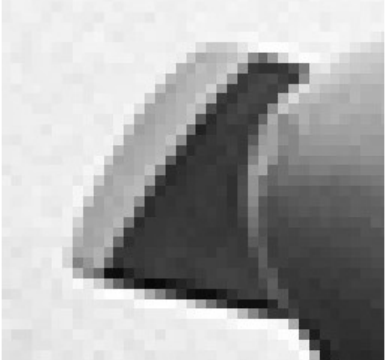} \\\vspace {0.5mm}
  \includegraphics[width=1.15\textwidth]{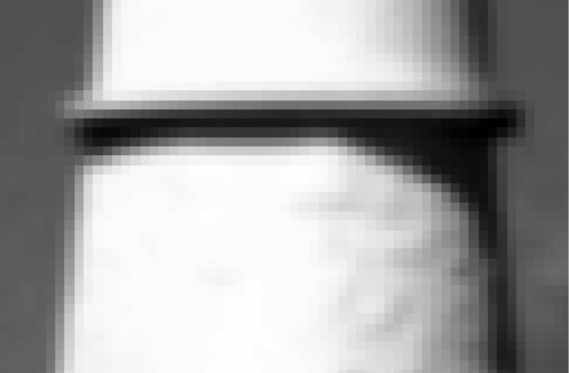} \\\vspace {0.5mm}
  \includegraphics[width=1.15\textwidth]{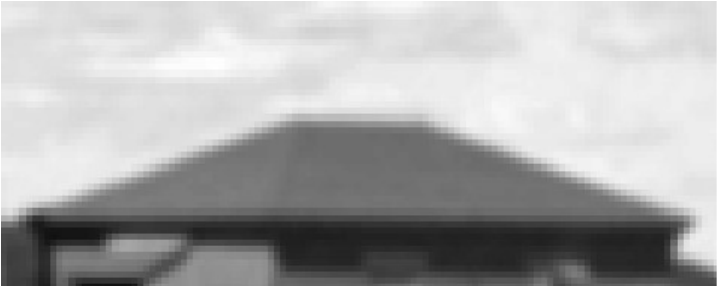} \\\vspace {0.5mm}
  \includegraphics[width=1.15\textwidth]{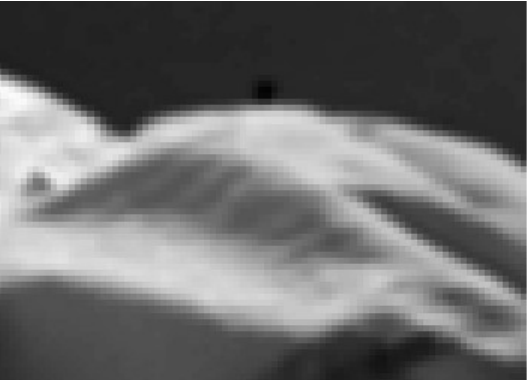} \\\vspace {0.5mm}
  \includegraphics[width=1.15\textwidth]{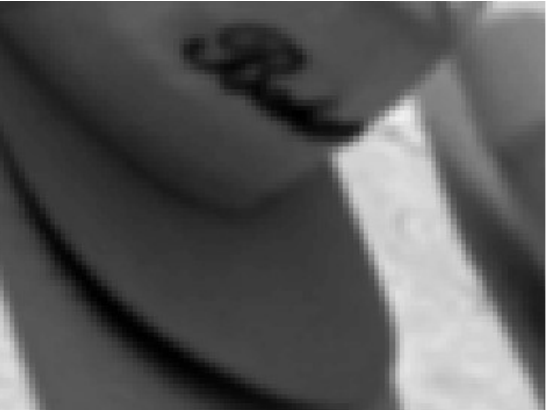} 
  \end{minipage}
}
\subfigure[]
{
  \begin{minipage}[b]{.11\linewidth}
  \centering
  \includegraphics[width=1.15\textwidth]{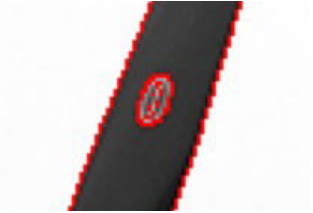} \\\vspace {0.5mm}
  \includegraphics[width=1.15\textwidth]{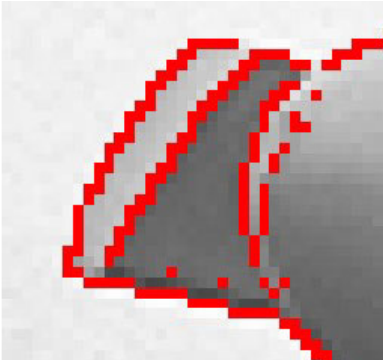} \\\vspace {0.5mm}
  \includegraphics[width=1.15\textwidth]{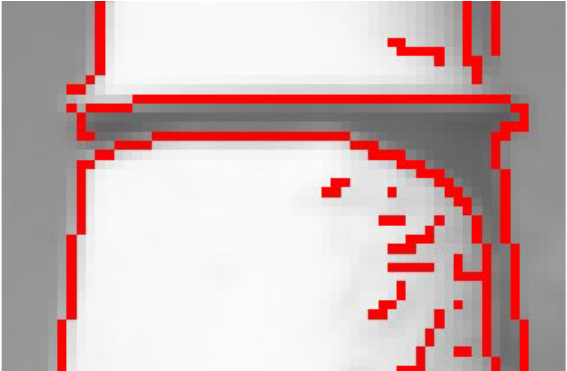} \\\vspace {0.5mm}
  \includegraphics[width=1.15\textwidth]{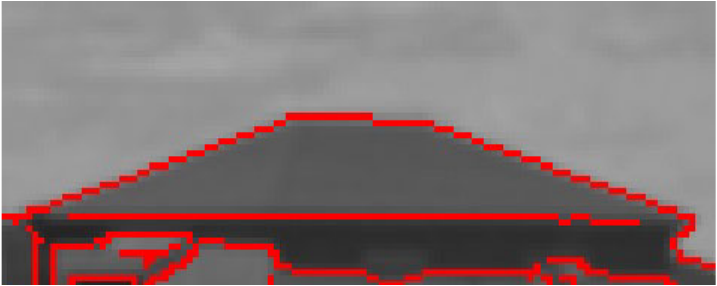} \\\vspace {0.5mm}
  \includegraphics[width=1.15\textwidth]{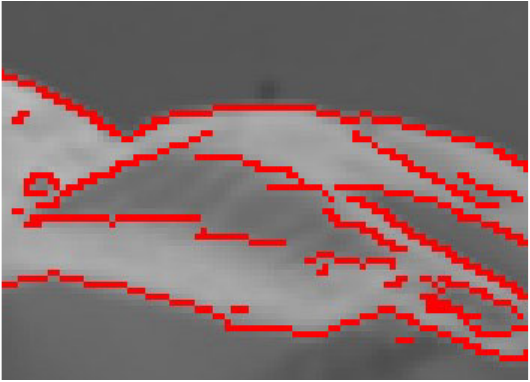} \\\vspace {0.5mm}
  \includegraphics[width=1.15\textwidth]{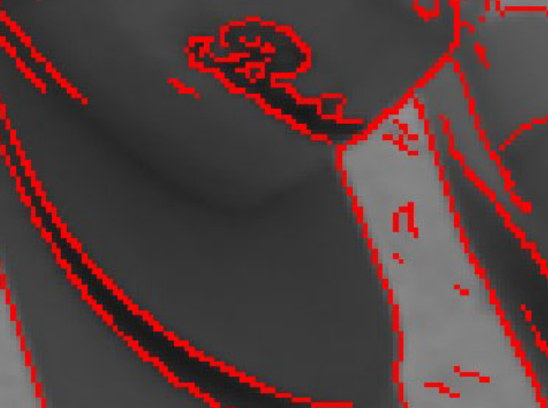} 
  \end{minipage}
}
\subfigure[]
{
  \begin{minipage}[b]{.11\linewidth}
  \centering
  \includegraphics[width=1.15\textwidth]{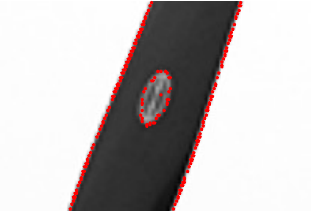} \\\vspace {0.5mm}
  \includegraphics[width=1.15\textwidth]{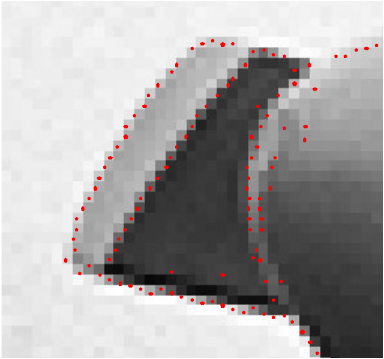} \\\vspace {0.5mm}
  \includegraphics[width=1.15\textwidth]{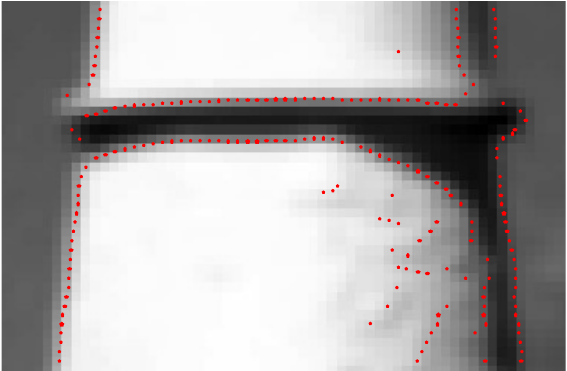} \\\vspace {0.5mm}
  \includegraphics[width=1.15\textwidth]{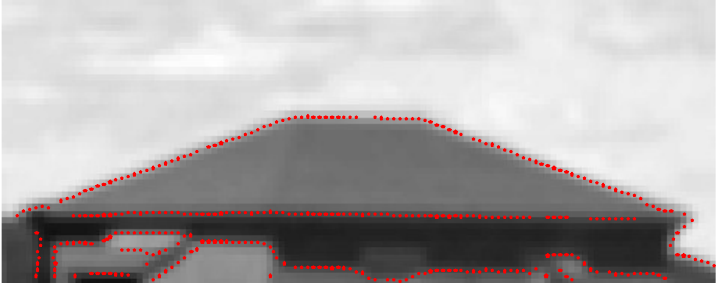} \\\vspace {0.5mm}
  \includegraphics[width=1.15\textwidth]{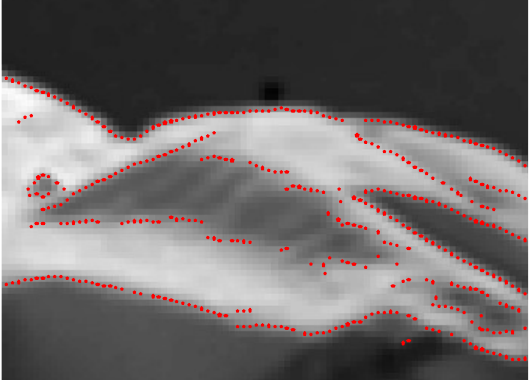} \\\vspace {0.5mm}
  \includegraphics[width=1.15\textwidth]{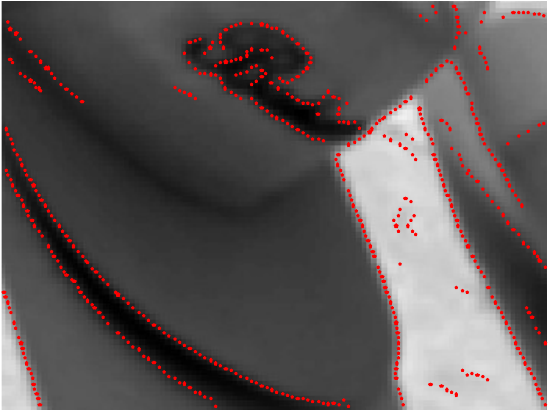} 
  \end{minipage}
}
\subfigure[]
{
  \begin{minipage}[b]{.11\linewidth}
  \centering
  \includegraphics[width=1.15\textwidth]{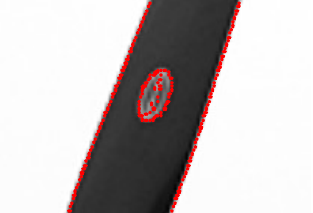} \\\vspace {0.5mm}
  \includegraphics[width=1.15\textwidth]{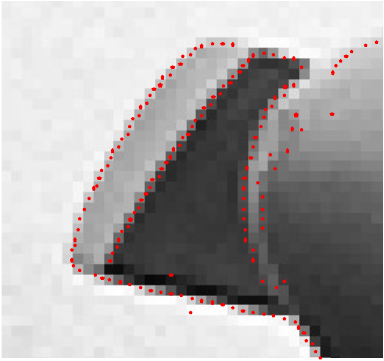} \\\vspace {0.5mm}
  \includegraphics[width=1.15\textwidth]{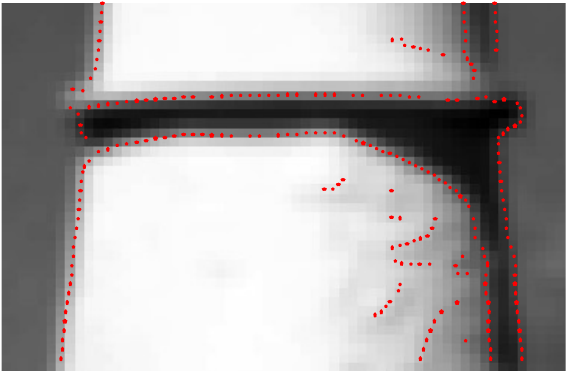} \\\vspace {0.5mm}
  \includegraphics[width=1.15\textwidth]{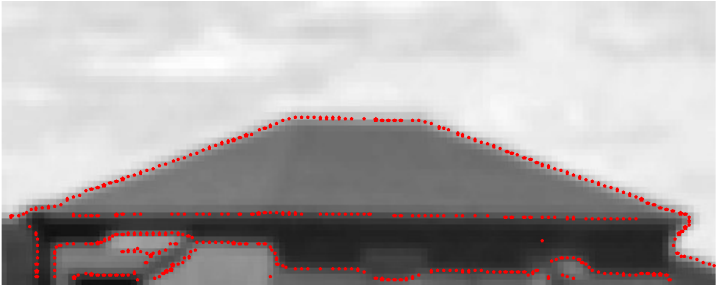} \\\vspace {0.5mm}
  \includegraphics[width=1.15\textwidth]{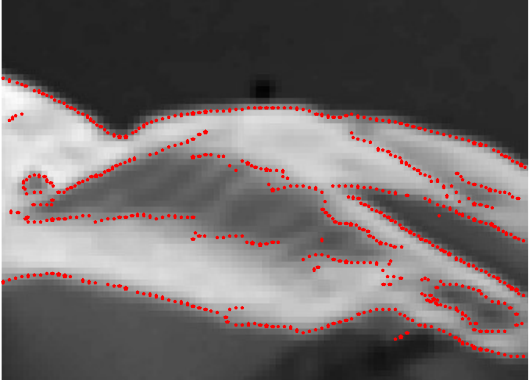} \\\vspace {0.5mm}
  \includegraphics[width=1.15\textwidth]{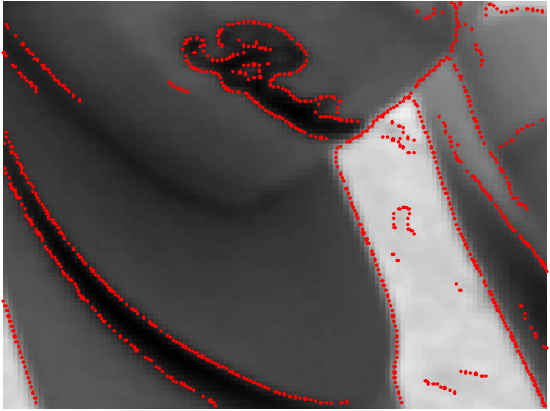} 
  \end{minipage}
}
\subfigure[]
{
  \begin{minipage}[b]{.11\linewidth}
  \centering
  \includegraphics[width=1.15\textwidth]{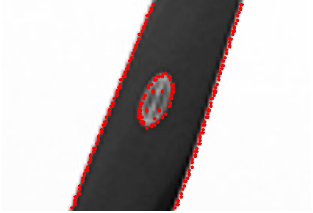} \\\vspace {0.5mm}
  \includegraphics[width=1.15\textwidth]{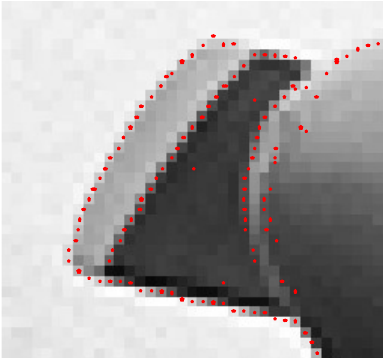} \\\vspace {0.5mm}
  \includegraphics[width=1.15\textwidth]{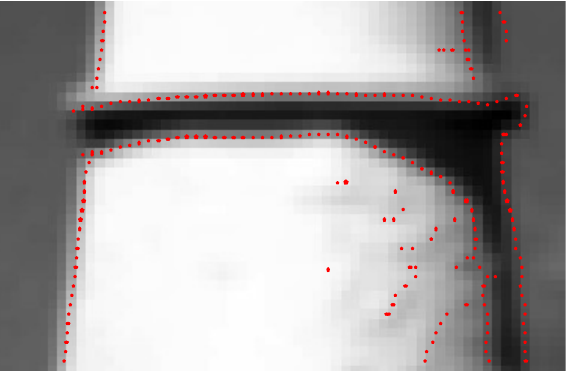} \\\vspace {0.5mm}
  \includegraphics[width=1.15\textwidth]{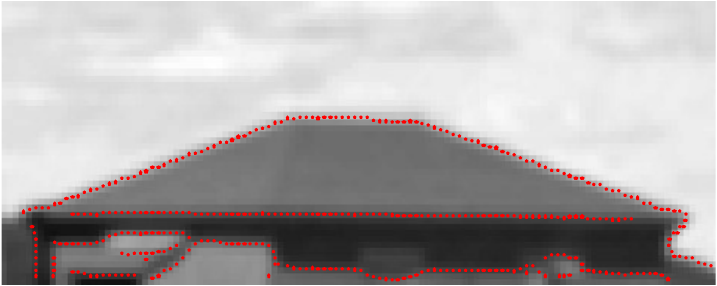} \\\vspace {0.5mm}
  \includegraphics[width=1.15\textwidth]{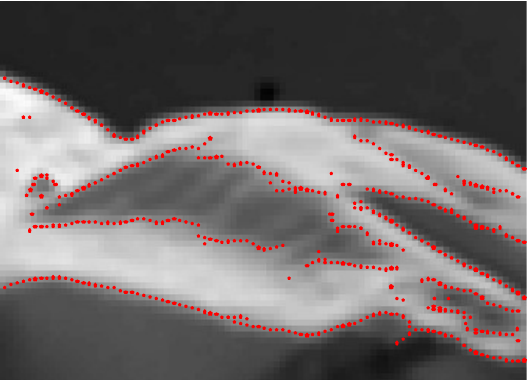} \\\vspace {0.5mm}
  \includegraphics[width=1.15\textwidth]{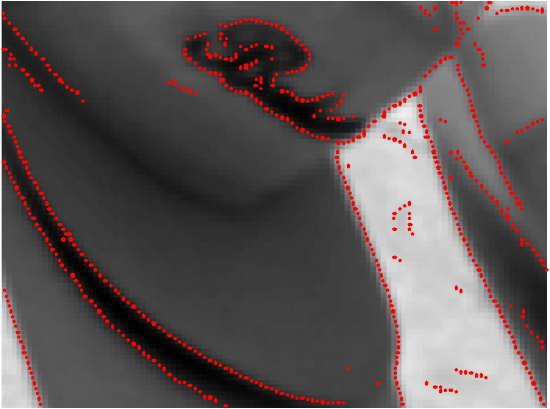} 
  \end{minipage}
}
\subfigure[]
{
  \begin{minipage}[b]{.11\linewidth}
  \centering
  \includegraphics[width=1.15\textwidth]{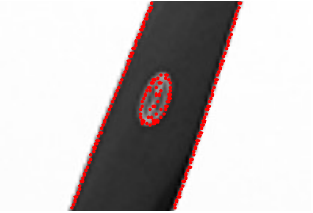} \\\vspace {0.5mm}
  \includegraphics[width=1.15\textwidth]{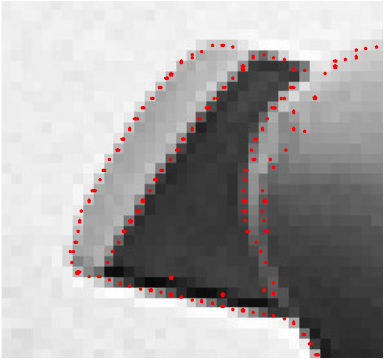} \\\vspace {0.5mm}
  \includegraphics[width=1.15\textwidth]{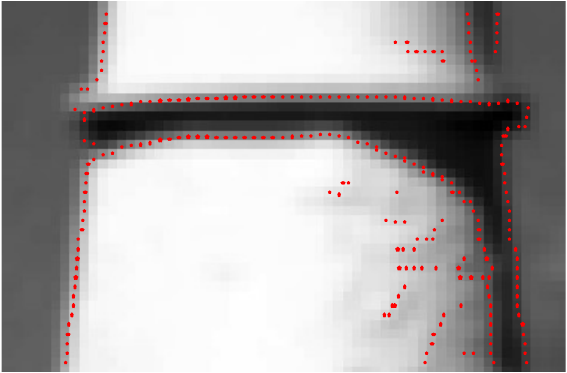} \\\vspace {0.5mm}
  \includegraphics[width=1.15\textwidth]{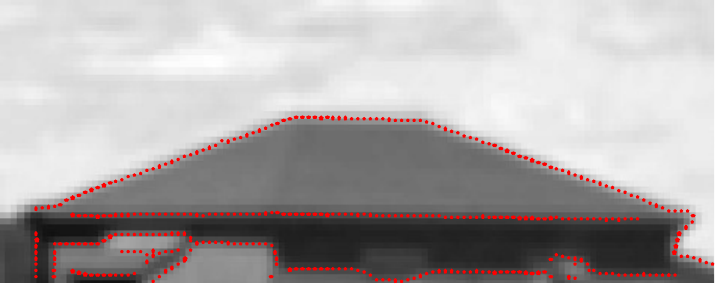} \\\vspace {0.5mm}
  \includegraphics[width=1.15\textwidth]{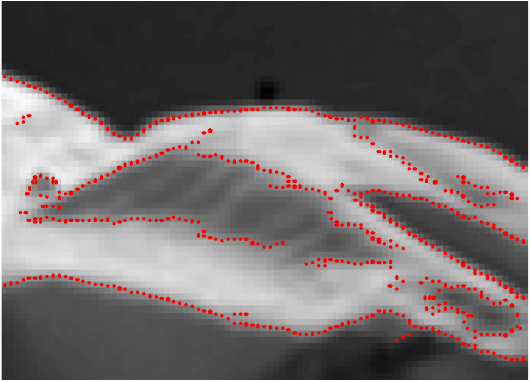} \\\vspace {0.5mm}
  \includegraphics[width=1.15\textwidth]{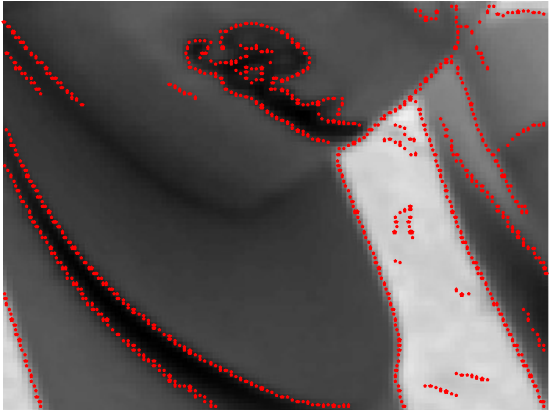} 
  \end{minipage}
}
\subfigure[]
{
  \begin{minipage}[b]{.11\linewidth}
  \centering
  \includegraphics[width=1.15\textwidth]{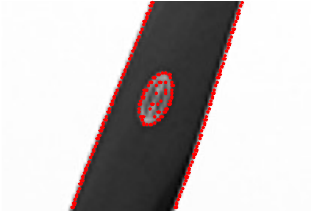} \\\vspace {0.5mm}
  \includegraphics[width=1.15\textwidth]{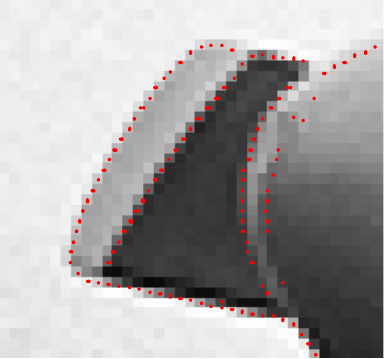} \\\vspace {0.5mm}
  \includegraphics[width=1.15\textwidth]{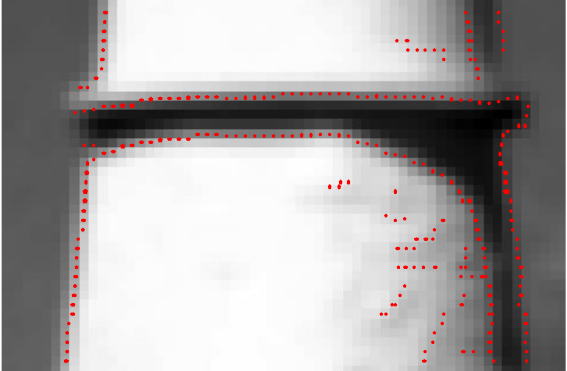} \\\vspace {0.5mm}
  \includegraphics[width=1.15\textwidth]{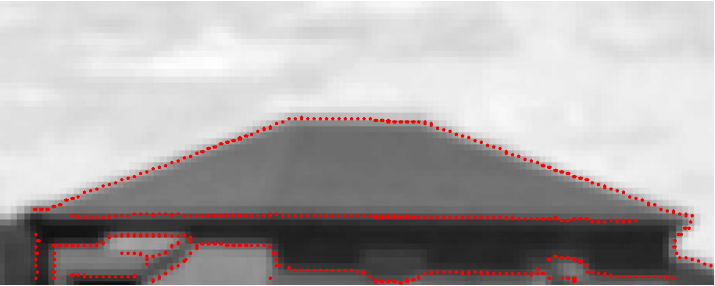} \\\vspace {0.5mm}
  \includegraphics[width=1.15\textwidth]{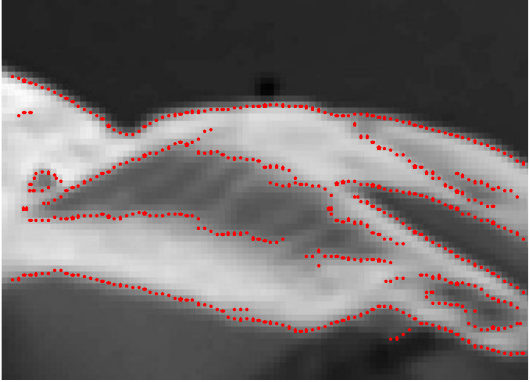} \\\vspace {0.5mm}
  \includegraphics[width=1.15\textwidth]{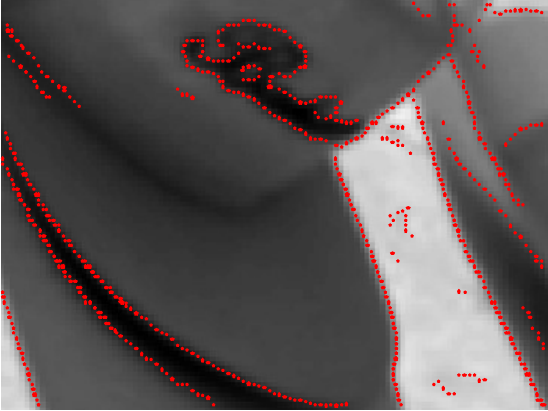} 
  \end{minipage}
}
\caption{Subpixel localization results on the sub-images in Fig. \ref{fig16}. (a) Original image. (b) Pixel level edges detected by Canny. (c) PAE. (d) Polynomial interpolation. (e) Erf. (f) CIS. (g) CIS+SER.}
\label{fig17}
\end{figure*}

\begin{figure}[t!]
\centering
\subfigure[]{
    \begin{minipage}[b]{.15\linewidth}
     \centering
        \includegraphics[width= 0.9in]{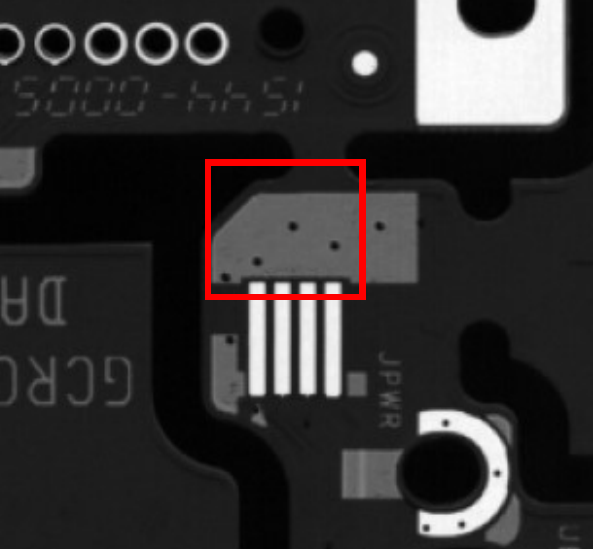}\\
        \vspace {0.1cm}
        \includegraphics[width= 0.9in]{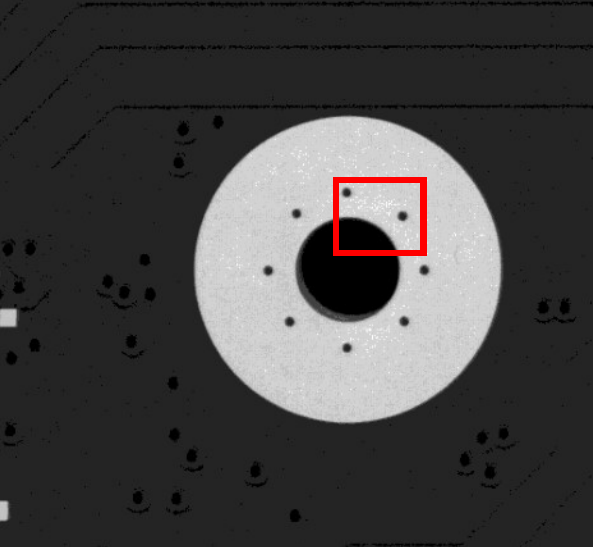}\\
        \vspace {0.1cm}
        \includegraphics[width= 0.9in]{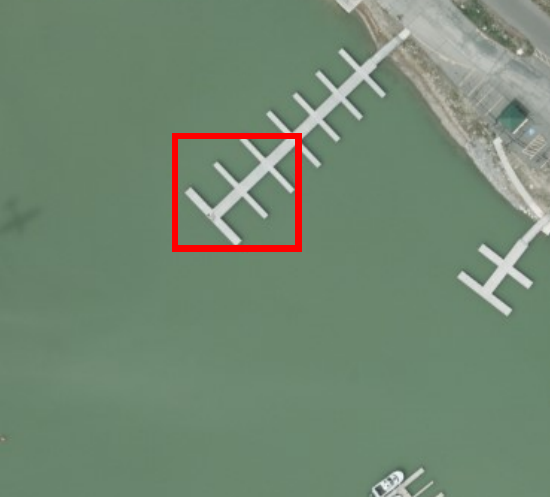} \\
        \vspace {0.1cm}
        \includegraphics[width= 0.9in]{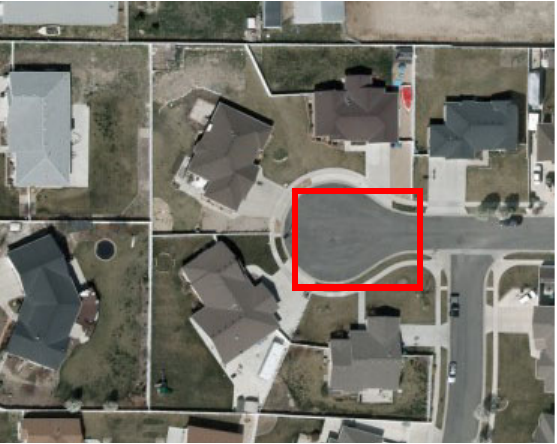} \\
    \end{minipage}
    }
\subfigure[]{
    \begin{minipage}[b]{.15\linewidth}
     \centering
        \includegraphics[width= 0.9in]{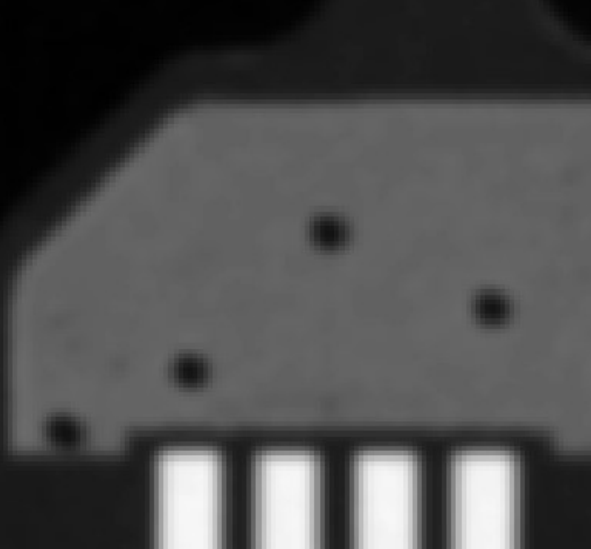}\\
        \vspace {0.1cm}
        \includegraphics[width= 0.9in]{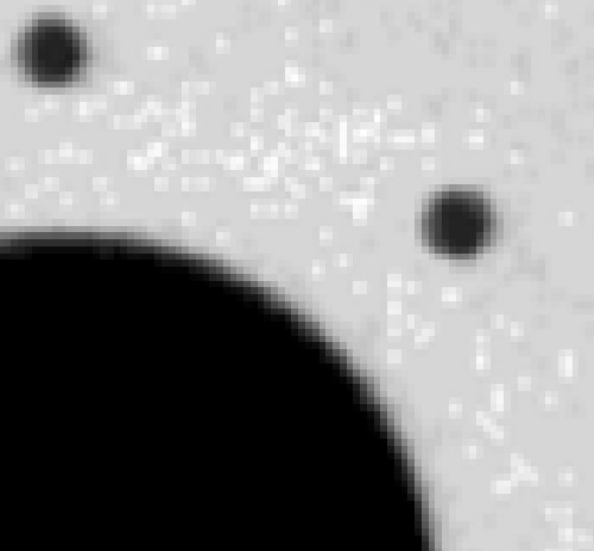}\\
        \vspace {0.1cm}
        \includegraphics[width= 0.9in]{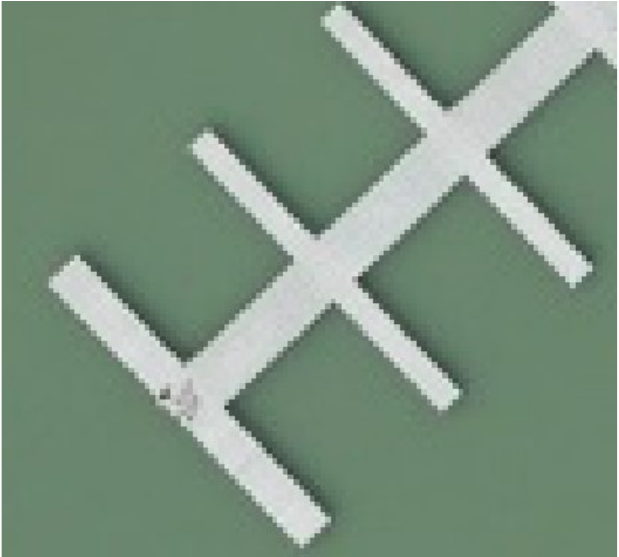} \\
        \vspace {0.1cm}
        \includegraphics[width= 0.9in]{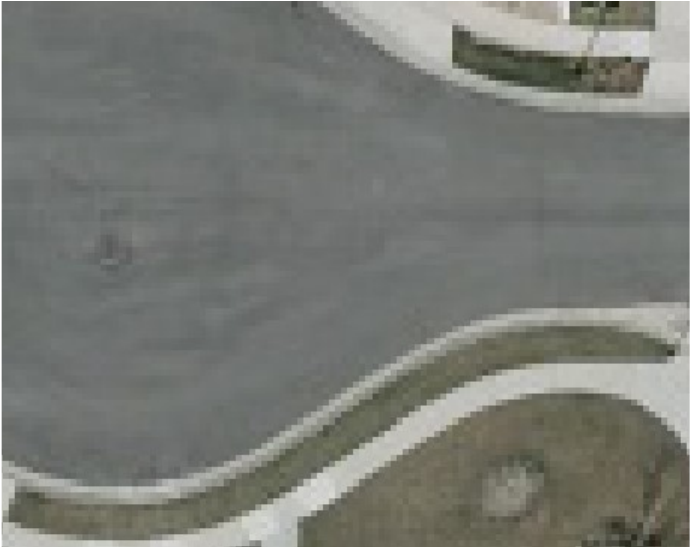} \\
    \end{minipage}
    }
\subfigure[]{
    \begin{minipage}[b]{.15\linewidth}
     \centering
        \includegraphics[width= 0.9in]{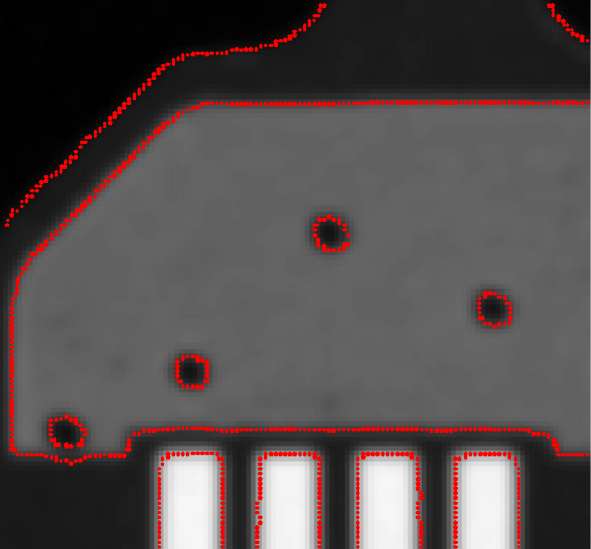}\\
        \vspace {0.1cm}
        \includegraphics[width= 0.9in]{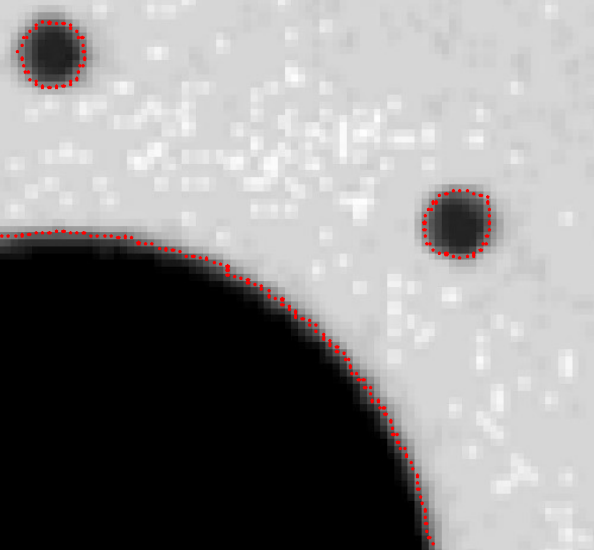}\\
        \vspace {0.1cm}
        \includegraphics[width= 0.9in]{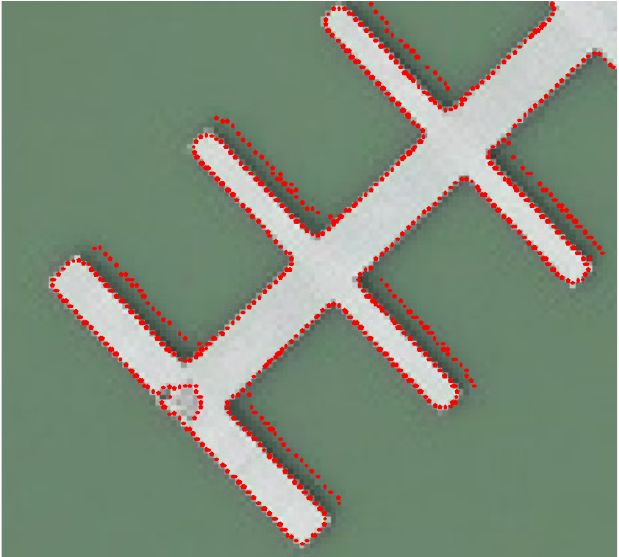} \\
        \vspace {0.1cm}
        \includegraphics[width= 0.9in]{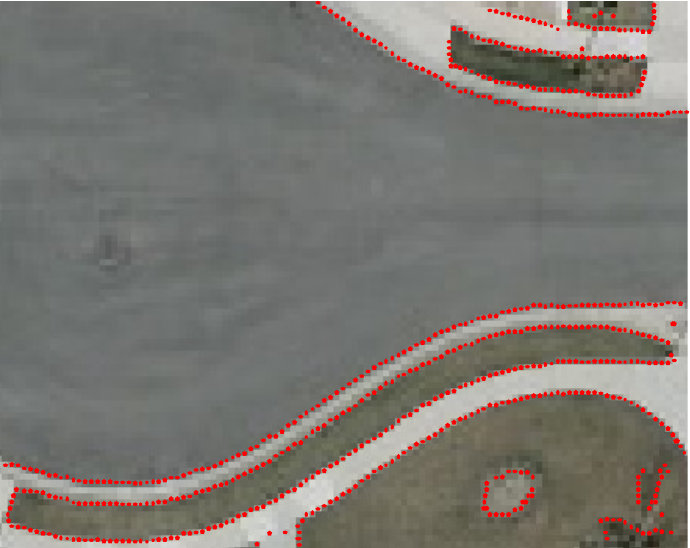} \\
    \end{minipage}
    }
\subfigure[]{
    \begin{minipage}[b]{.15\linewidth}
     \centering
        \includegraphics[width=0.838in]{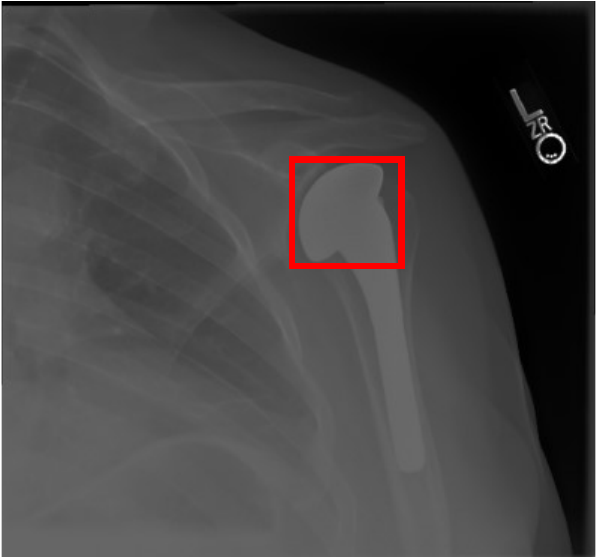} \\
        \vspace {0.1cm}
        \includegraphics[width=0.838in]{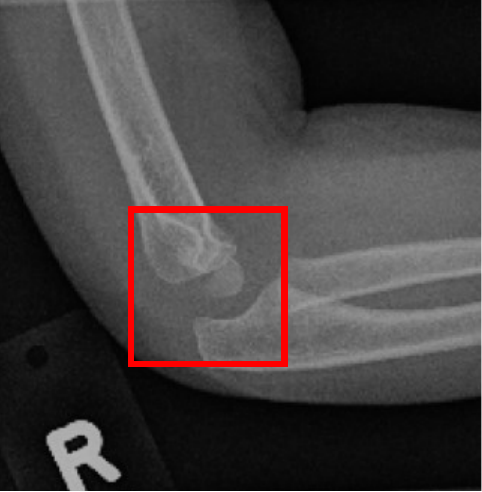}\\
        \vspace {0.1cm}
        \includegraphics[width=0.838in]{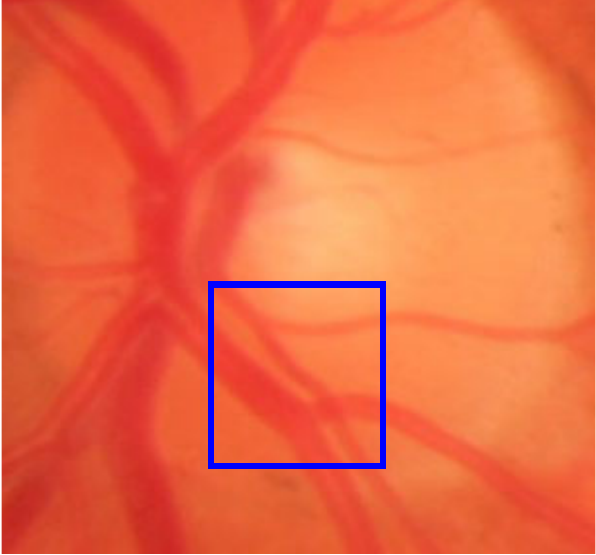} \\
        \vspace {0.1cm}
        \includegraphics[width=0.838in]{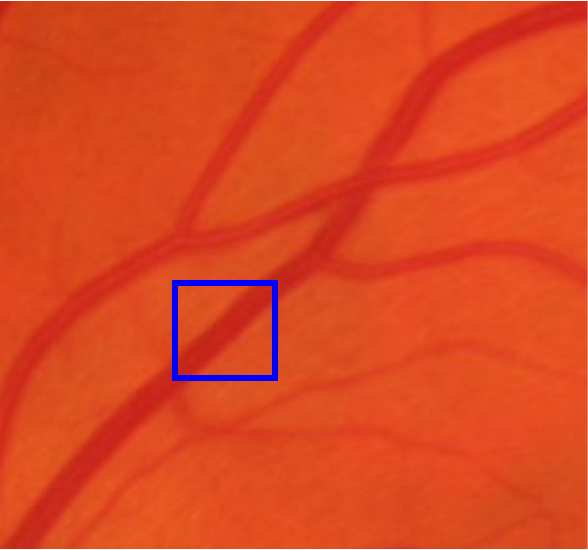}
    \end{minipage}
    }
\subfigure[]{
    \begin{minipage}[b]{.15\linewidth}
     \centering
        \includegraphics[width=0.838in]{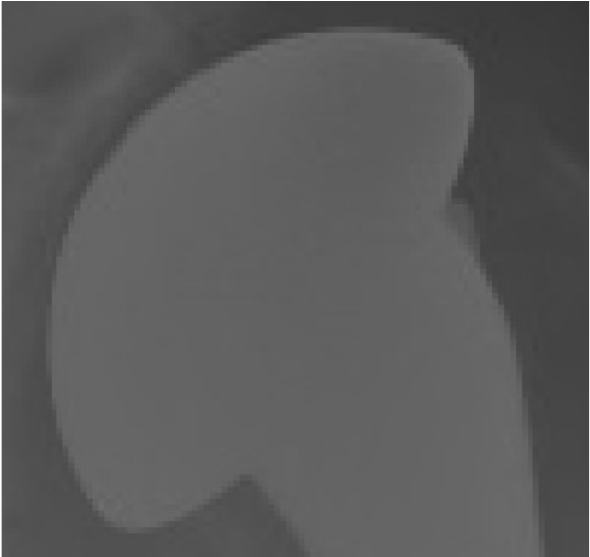} \\
        \vspace {0.1cm}
        \includegraphics[width=0.838in]{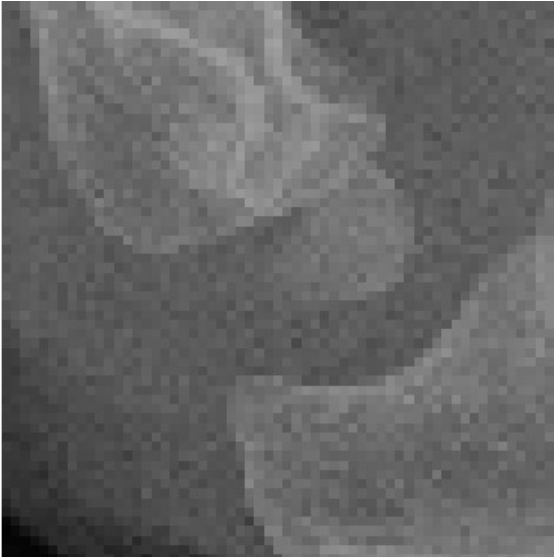}\\
        \vspace {0.1cm}
        \includegraphics[width=0.838in]{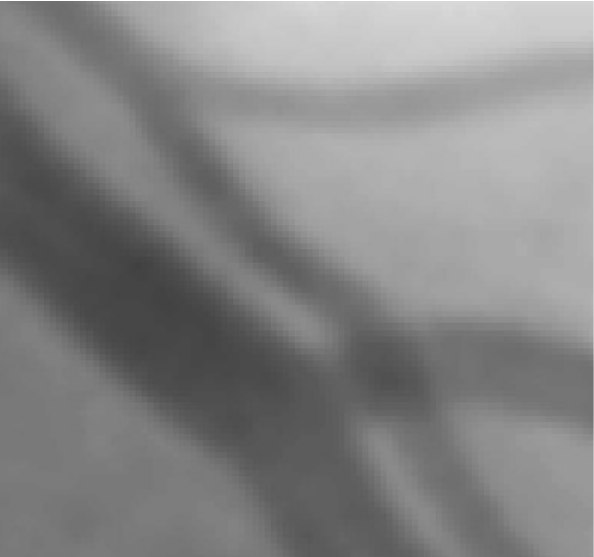} \\
        \vspace {0.1cm}
        \includegraphics[width=0.838in]{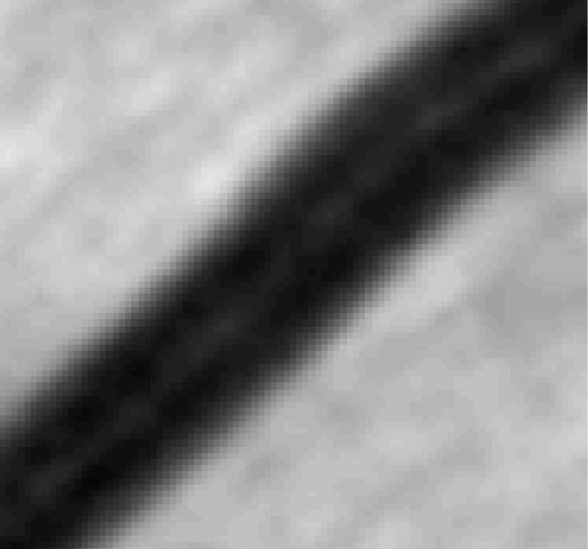}
    \end{minipage}
    }
\subfigure[]{
    \begin{minipage}[b]{.15\linewidth}
     \centering
        \includegraphics[width=0.838in]{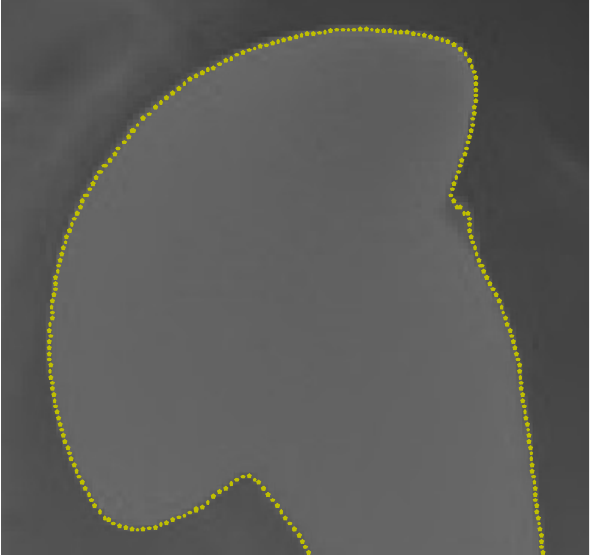} \\
        \vspace {0.1cm}
        \includegraphics[width=0.838in]{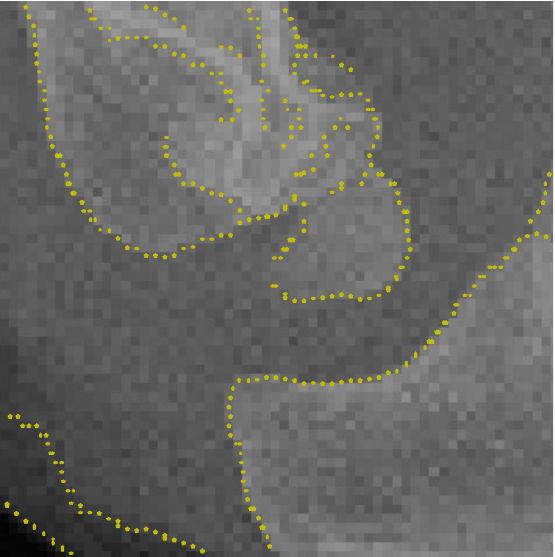}\\
        \vspace {0.1cm}
        \includegraphics[width=0.838in]{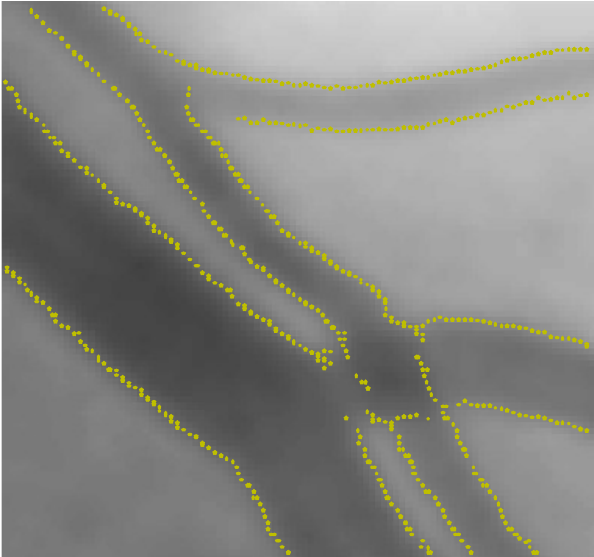} \\
        \vspace {0.1cm}
        \includegraphics[width=0.838in]{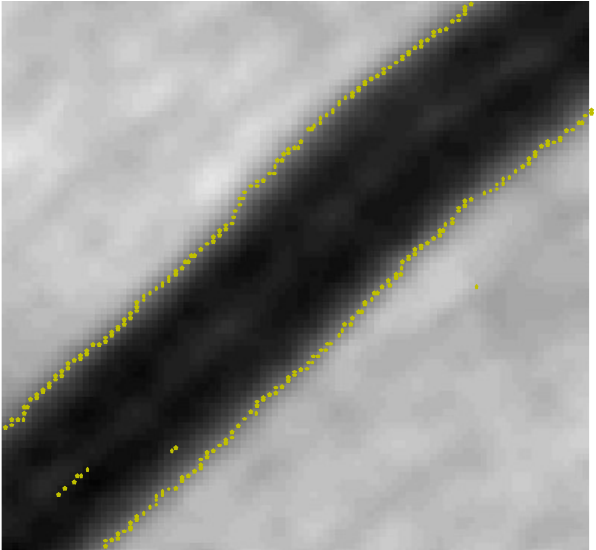}
    \end{minipage}
    }
\caption{Subpixel edge Localization results on PCB, remote sensing images, and medical images. (a) Original image of PCB, remote sensing images. (b) Enlarged sub-images bounded by red rectangle in (a). (c) Subpixel localization results on (b).  (d) Original image. The two rows on the top are human joints and the two rows on the bottom are capillaries in the eyeball. (e) Enlarged sub-image in (d). (f) Subpixel edge localization results on (e).}
\label{fig19}
\end{figure}

\subsubsection{Analysis of slant dataset}
The size of samples is 221 × 221, with lines drawn from horizontal coordinates of 10 to 200 at a interval of 10 pixels. The intensity of the side below the lines is set to 200 while that on the opposite side is 50. 
The $SNR$ remains 85 in this dataset, while $k_G$ takes the values of 5 and 7.
The slope varies from 1 to 10, the sharper the angle formed by two lines, the greater the mutual interference between them.
Similarly, each set includes 5 samples, and the RMSE is calculated between the actual and measured vertical coordinates of each edge point.

As demonstrated by Fig. \ref{fig15}, the proposed method achieves relatively optimal accuracy under various slopes and filtering radius. 
CIS exhibits strong resistance to mutual interference and blurring. The error grows by 0.2 pixels as the slope increases by 1 from 1 to 4, and grows by 0.3 pixels as the slope increases by 1 from 5 to 10.
The SER can reduce the error of CIS by up to 0.25 pixels, but its efficacy is constrained under specific conditions. 
Specifically, the intensity distribution within the region becomes disordered at high noise levels. This disruption may impair accurate parameter extraction, diminish the beneficial effect of SER as well.
However, in most scenarios, SER can effectively alleviate the negative impacts of the context. 

\begin{table}[t!]
\caption{average running time}
\centering
\label{tab4}
\setlength{\tabcolsep}{3.4pt}
\renewcommand\arraystretch{1.3}
\begin{tabular}{ccccccc}
\hline
Method & Interpolation & Erf & PAE & CIS & CIS+SER\\ 
\hline
Mean time (sec.) & 0.1531 & 0.89393 & 0.0365 & 0.02937 & 0.37211\\ 
\hline
\end{tabular}
\end{table}

\subsubsection{Running Time}
The average running time of the five methods on the circle dataset is shown in Table \ref{tab4}. Notably, CIS exhibits superior performance of taking mere 0.03 sec. in processing one image. Despite the relatively weak anti-interference ability, it still maintains comparable performance to other methods, as illustrated in Fig. \ref{fig13}.
Unfortunately, the incorporation of SER with CIS leads to a remarkable increase in running time. However, even at tenfold time cost, the significant accuracy improvement justifies this trade-off.

In conclusion, CIS is the most cost-effective method, offering good balance in terms of time and accuracy. The integration of SER into CIS can attain the highest precision by acquiring more robust parameters at the cost of more time consumption.

\subsection{Experiments on real images}\label{sec5.2}
In this section, we offer the qualitative analysis in real images without the ground truth and provide the subpixel edge localization results for various applications.

\subsubsection{Qualitative analysis}
Several images from TID2013 \cite{Ponomarenko2015ImageDT} dataset are chosen for qualitative analysis, as depicted in Fig. \ref{fig16}. The sub-images bounded by the red rectangles are enlarged to facilitate the analysis. Moreover, Canny is utilized to determine the initial localization of edge pixels.

Fig. \ref{fig17} displays the positioning results. Compared to the other algorithms exhibiting obvious errors, such as interrupted edges and outliers, CIS is better in most cases.
However, due to the susceptibility to the interference resulting from uneven local intensity distribution, CIS yields the disordered outcomes at the central logos in Image 1 and 6. The introduction of SER, effectively mitigates deviations by disregarding the impact of local interference and provides a cleaner context, as illustrated in Image 1, 2 and 5. But the compound method may cause slight invalid offsets, thereby destructing the smoothness of the curve to some extent, as depicted in Image 3 and 6.

\subsubsection{Analysis of real images from various applications}
The proposed compound method is finally qualitatively evaluated using real-world images from diverse fields, including industrial PCB images, remote sensing images and medical images. 

Fig. \ref{fig19}(a-c) provides the test results of the proposed overall framework on industrial PCB and remote sensing images. As expected, the method exhibits the robust resistance to common interferences in industrial image such as noise and blurring, as well as the excellent capability of accurately locating the edge details in remote sensing images. 

Test results on medical images of joints and capillaries are shown in Fig. \ref{fig19}(d-f). For the first row image, the algorithm works well due to the clean boundary. In contrast, the second row is a common fuzzy image resulting from insufficient equipment accuracy or field shift effect, but the edge localization result is overall satisfactory. In the capillary image, the extracted edges may not be perfectly smooth, but roughly reflect the changes of capillary edges. It is also noted that in some intricate regions such as the internal structures of bones in the second row image or the intersections of capillaries in the third row image, the traditional method considering intensity difference of edges may fail. In this case, it may be a better solution to identify the initial edge outline through deep learning models with semantic information and apply the proposed method to determine the optimal subpixel edge locations. 

\section{Conclusion}\label{sec6}
In this paper, we propose a straightforward and effective subpixel edge localization method based on converted intensity summation. 
To enhance the anti-interference capability of the method, stable edge region is proposed to capture the robust statistical parameter. And an edge complement method based extension-adjustment is suggested for irregular edges correction.
Experiments on simulated images have proven that CIS can achieve superior positioning accuracy compared to the current state-of-the-art method while maintaining the fastest running speed. Although the incorporation of the SER increases the time cost, it observably improves the localization performance. The experiments on real-world images have verified the effectiveness of the proposed method, demonstrating its suitability in applications requiring high-precision of edge localization. Future work will involve the optimization of the compound algorithm aiming to specific application scenarios.

\bibliographystyle{IEEEtran}  
\bibliography{citation}   
\end{document}